\documentclass{article}

\usepackage{microtype}
\usepackage{graphicx}
\usepackage{subfigure}
\usepackage{booktabs} 

\usepackage{hyperref}
\usepackage{multirow}
\usepackage{pifont}

\newcommand{\xmark}{\ding{55}}
\newcommand{\cmark}{\ding{52}}

\usepackage[accepted]{icml2025}

\usepackage{amsmath}
\usepackage{amssymb}
\usepackage{mathtools}
\usepackage{amsthm}

\usepackage[capitalize,noabbrev]{cleveref}

\theoremstyle{plain}

\theoremstyle{definition}

\theoremstyle{remark}

\usepackage[textsize=tiny]{todonotes}

\icmltitlerunning{APC-VFL}

\begin{document}

\twocolumn[
\icmltitle{Towards Active Participant Centric Vertical Federated Learning: Some Representations May Be All You Need}




\begin{icmlauthorlist}
\icmlauthor{Jon Irureta}{iker}
\icmlauthor{Jon Imaz}{iker}
\icmlauthor{Aizea Lojo}{iker}
\icmlauthor{Javier Fernandez-Marques}{flwr}
\icmlauthor{Iñigo Perona}{ehu}
\icmlauthor{Marco González}{iker}

\end{icmlauthorlist}

\icmlaffiliation{iker}{Ikerlan Technology Research Centre, Arrasate, Spain,}
\icmlaffiliation{flwr}{Flower Labs, Cambridge, UK,}
\icmlaffiliation{ehu}{University of the Basque Country UPV/EHU, Donostia, Spain}

\icmlcorrespondingauthor{Jon Irureta}{jirureta@ikerlan.es}

\icmlkeywords{Machine Learning, ICML}

\vskip 0.3in
]



\printAffiliationsAndNotice{}  

\begin{abstract}
Existing Vertical FL (VFL) methods often struggle with realistic and unaligned data partitions, and incur into high communication costs and significant operational complexity. This work introduces a novel approach to VFL, \textit{Active Participant Centric VFL (APC-VFL)}, that excels in scenarios when data samples among participants are partially aligned at training. Among its strengths, APC-VFL only requires a single communication step with the active participant. This is made possible through a local and unsupervised representation learning stage at each participant followed by a knowledge distillation step in the active participant. Compared to other VFL methods such as SplitNN or VFedTrans, APC-VFL consistently outperforms them across three popular VFL datasets in terms of F1, accuracy and communication costs as the ratio of aligned data is reduced.
\end{abstract}

\section{Introduction}
Federated learning (FL) is a distributed computational paradigm whose key aspect is that it allows collaborative modeling training while preserving data privacy \cite{kairouz2021advances, li2020federated}. As a result, FL is particularly relevant in fields such as medicine or finance, where data protection is a must \cite{LIU2025128653, long2020federated}.  

Federated learning can be split into two subgroups based on the data partition: Horizontal FL (HFL) and Vertical FL (VFL). The first case is the one where most research has been done, and assumes that participants have a common feature space but different sample space, which for instance, has allowed to collaboratively training a cancer boundary detection model, leveraging data from across the world \cite{pati2022federated}.
On the other hand, in VFL, the participants have a common sample space but differ in feature space. The federation process allows them to have access to more features, more information of samples, being one of the use cases training recommender systems \cite{tan2020federated}. It must be mentioned that under this setting a single participant, known as active participant, has access to labels, with the rest being, passive participants \cite{liu2024vertical}.

VFL faces two major issues, the communication cost and the data partition itself \cite{wei2022vertical}. In terms of communication, classical VFL requires passive participants to communicate with the active one on each iteration. As no local training is performed this often results in a high communication cost \cite{castiglia2022compressed, 9855231}. In the data partition side, it has traditionally been assumed that all samples are aligned among participants, i.e. all samples are present in all participants of the federation process. This assumption is rather naive, as it implies that all collaborators in the federated learning process, such as hospitals or banks, would have an identical set of patients or clients. Consequently, in real-world scenarios, the overlap between samples is often limited to a fraction of the overall dataset, presenting a significant challenge for vertical federated learning. \cite{Kang_2022, he2024hybrid}.  

APC-VFL aims to tackle both issues simultaneously: high communication and limited alignment ratios. With this goal, we propose to combine local unsupervised representation learning based on autoencoders \cite{tschannen2018recent} with knowledge distillation \cite{Gou_2021}. We present:

\begin{itemize}
    \item A novel VFL method that requires a single communication round and that is not limited to the usage of aligned samples for training or inference. This results in average reductions of up to $634\times$ communication rounds on MIMIC-III, $380\times$ in Breast Cancer Wisconsin and $1590\times$ in UCI Credit Card compared to SplitNN and without model degradation.
    \item A thorough evaluation process demonstrating that APC-CFL consistently outperforms a comparable state-of-the-art approach, VFedTrans \cite{huang2023vertical}, in multiple scenarios with limited sample overlap, reducing required communication rounds and system complexity. We reduce the communication footprint by up to $78.24\times$ on MIMIC-III, $2.20\times$ on Breast Cancer Wisconsin and $78.30\times$ on UCI Credit Card while obtaining improvements of more than $5$ points in F1 score, more than $2\%$ in accuracy and more than $1$ point in F1 score on the respective datasets.
\end{itemize}


\section{Background and related work}
\subsection{Vertical FL with complete sample alignment}
The goal of VFL is to train a model in a privacy preserving manner while having the feature space divided between different participants \cite{li2020federated}.

Assuming that all samples are aligned among participants, several approaches have been proposed to train tree-based models, such as SecureBoost \cite{cheng2021secureboost}, its variant SecureBoost+ \cite{chen2021secureboost+}, and Federated Forest \cite{liu2020federated}.

\begin{figure}[h]
    \centering
    \subfigure[Forward pass.]{
        \includegraphics[width=0.20\textwidth]{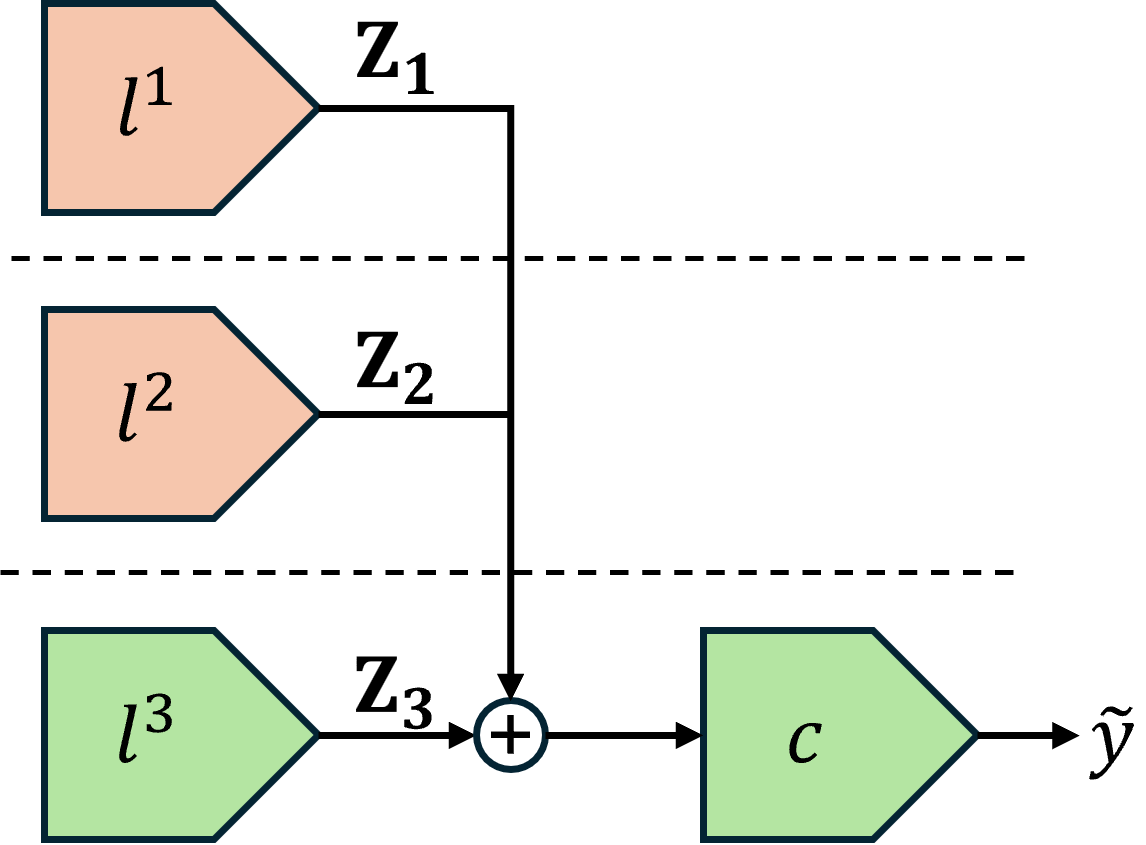}
        \label{subfig:forward}
    }
    \hfill
    \subfigure[Backpropagation.]{
        \includegraphics[width=0.20\textwidth]{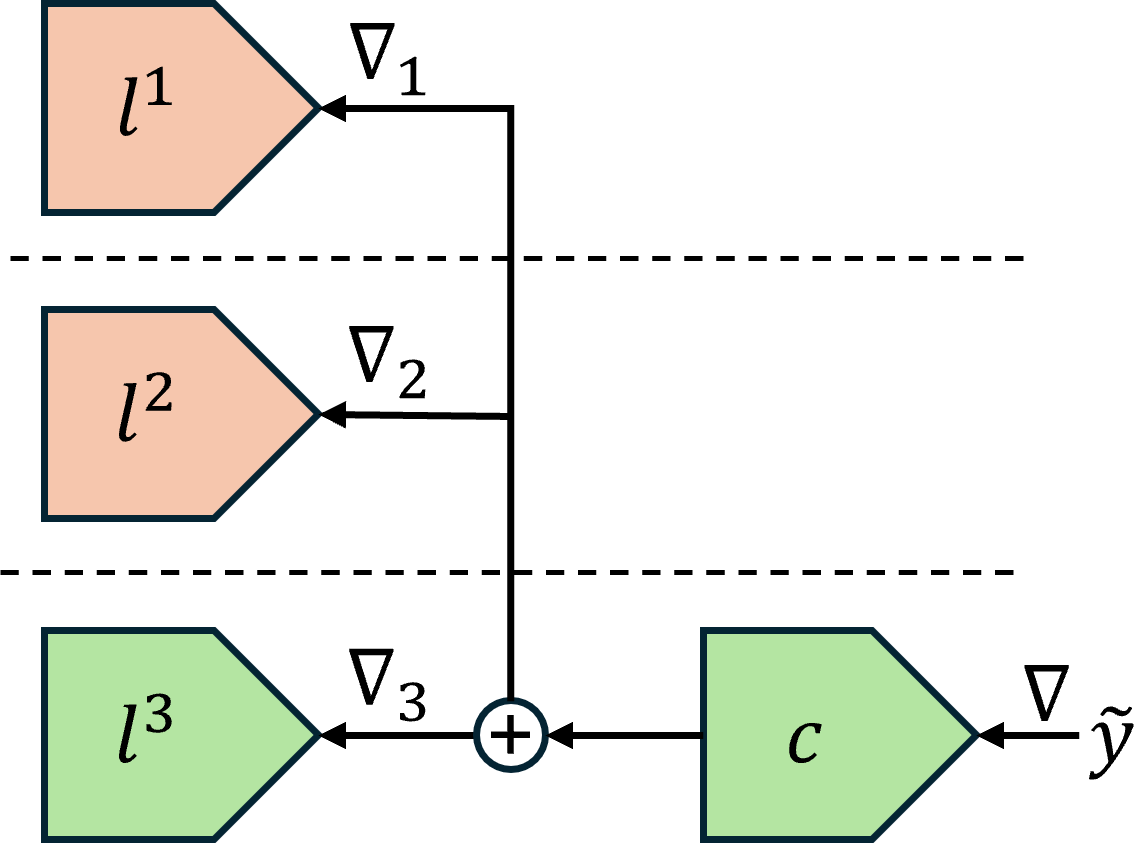}
        \label{subfig:back}
    }
    \caption{Representation of forward pass (\ref{subfig:forward}) and backpropagation (\ref{subfig:back}) on a Vertical SplitNN scenario with three participants.}
    \label{fig:splitNN_forward_backward}
\end{figure}

Under the same assumption, Vertical SplitNN adapts neural networks to the VFL setting by means of split learning \cite{vepakomma2018split, ceballos2020splitnndriven}, which is a common VFL baseline. In Vertical SplitNN, each participant of the federation process has a local feature extractor, $l^k$, and the active party has a classification head as well, $c$, as represented in Figure \ref{fig:splitNN_forward_backward}. In an ideal scenario, each participant $k$ performs a forward pass of a batch using $l^k$ and sends the resulting embeddings, $Z_k$, to the active party. The active party concatenates these embeddings and performs the classification using $c$ (see Figure \ref{subfig:forward}). Once the classification is done, the classification head is updated with regular gradient descent and the active participant sends the gradients that correspond to the $k^{th}$ local model to the corresponding participant to update it (see Figure \ref{subfig:back}).

It is important to note that while concatenation is the most effective strategy for embedding aggregation, it suffers the greatest performance degradation when participants encounter communication issues, e.g. a client drops from the federation \cite{li2023fedvs}.

\subsection{Vertical FL with limited sample alignment}
\begin{figure}[!h]
    \centering
    \includegraphics[width=0.70\columnwidth]{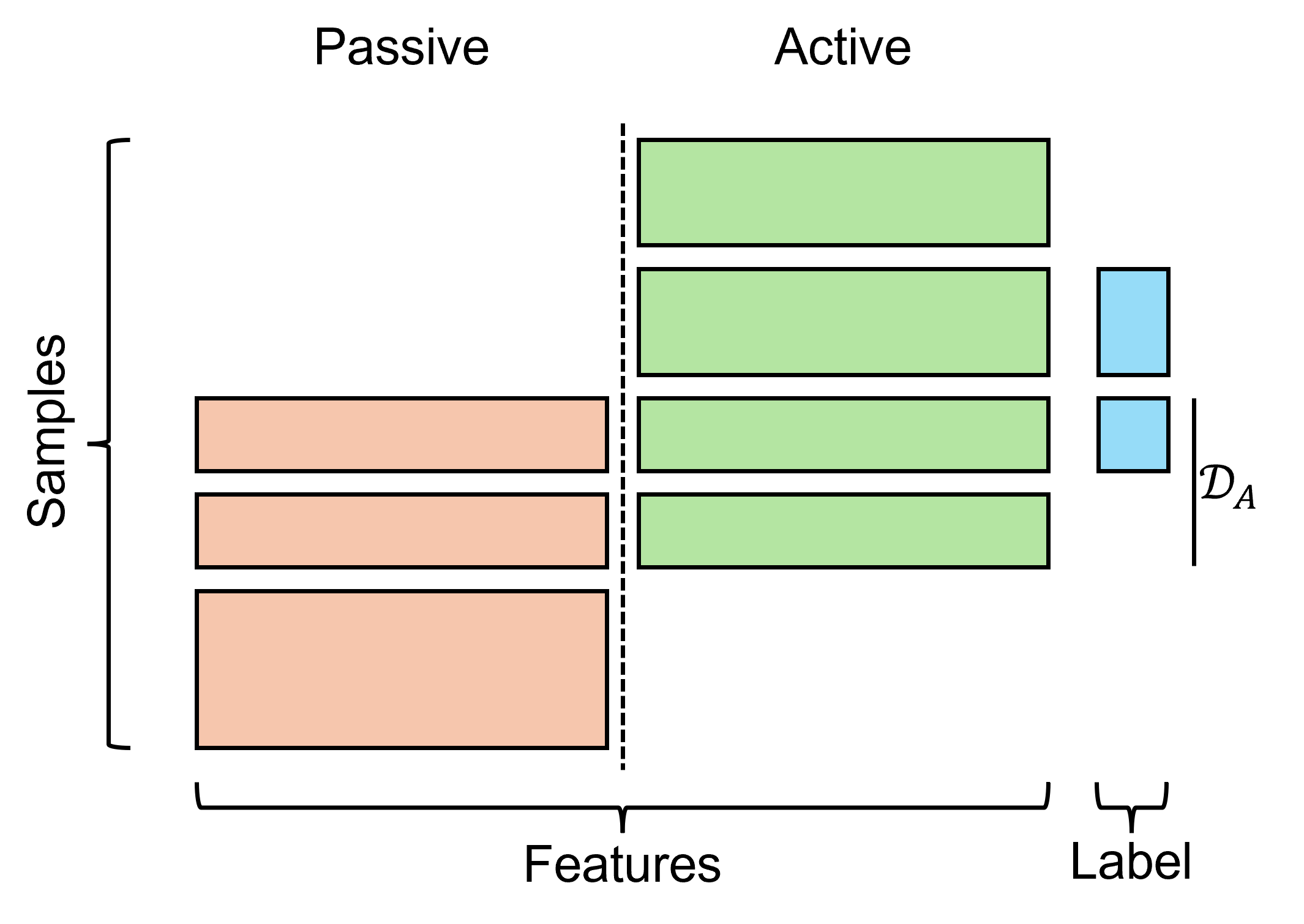}
    \caption{Data partition in VFL with two participants. $\mathcal{D}_A$, the set of aligned samples, is a small fraction of the whole dataset.}
    \label{fig: data VFL}
\end{figure}

When not all samples are aligned, as represented in Figure \ref{fig: data VFL}, a more complex and real scenario is presented, for which different approaches have been proposed in the literature. FedHSSL \cite{he2024hybrid} tries to tackle the problem using self-supervised learning via Siamese Networks \cite{chicco2021siamese}. Similarly, FedCVT \cite{Kang_2022} leverages self-supervised learning and scaled dot-product attention \cite{vaswani2017attention} to estimate the missing representations of unaligned data, increasing the data each participant has available. 


While algorithmically interesting, these approaches are still restricted to the aligned samples during inference and the required information exchange and synchronization among participants makes them challenging to implement in practical, real-world, settings. To address this, One-Shot/Few-shot VFL \cite{sun2023communication} does take into account the communication overheard, achieving good results with only three communication rounds by combining clustering and self-supervised learning. Even if this work takes an important step towards making VFL \textit{``real-world ready''}, there still exists a big gap in the literature: the usage of unaligned samples during inference, in an independent manner, by the active participant.

The inability to leverage unaligned samples for inference severely limits the applicability of recent VFL frameworks. VFedTrans \cite{huang2023vertical} addresses this while relying on FedSVD \cite{chai2022practical} or VFedPCA \cite{cheung2022vertical} to learn representations of the aligned samples, which are then used in a knowledge distillation process. In order to learn those aligned representations, multiple communication rounds and a third party server are required, which increases the complexity for VFedTrans to be the deployed in the real world.

 With APC-VFL, which we detail in Section \ref{sec:method}, we present a framework that is not limited to the use of aligned samples for training or inference. It requires only a single exchange of information between participants and, can be implemented in peer-to-peer frameworks without the need for third party servers.

Table \ref{tab:vfl data usage and comm} summarizes the aforementioned proposals in terms of data usage, required information exchange and communication scheme.

\begin{table}[!h]
\scriptsize
\centering
\caption{Comparing the proposed \textit{APC-VFL}, with other VFL methods in terms of data usage, required information exchange between participants, and support for peer-to-peer (P2P) schemes.}
\vspace{3mm}
\scalebox{1.1}{
\begin{tabular}{lllll}
\hline
\textbf{Work} & \textbf{\begin{tabular}[c]{@{}l@{}}Train w.\\ unaligned\end{tabular}} & \textbf{\begin{tabular}[c]{@{}l@{}}Inference w.\\ unaligned\end{tabular}} & \textbf{\begin{tabular}[c]{@{}l@{}}Single info\\ exchange\end{tabular}} & \textbf{P2P} \\ \hline
\textit{SplitNN}       & \multicolumn{1}{c}{\xmark}                                                     & \multicolumn{1}{c}{\xmark}                                                          & \multicolumn{1}{c}{\xmark}                                               & \multicolumn{1}{c}{\cmark} \\ \hline
\textit{FedHSSL}       & \multicolumn{1}{c}{\cmark}                                                     & \multicolumn{1}{c}{\xmark}                                                          & \multicolumn{1}{c}{\xmark}                                               & \multicolumn{1}{c}{\xmark} \\ \hline
\textit{FedCVT}        & \multicolumn{1}{c}{\cmark}                                                     & \multicolumn{1}{c}{\xmark}                                                          & \multicolumn{1}{c}{\xmark}                                               & \multicolumn{1}{c}{\cmark} \\ \hline
\textit{One-shot VFL}  & \multicolumn{1}{c}{\cmark}                                                     & \multicolumn{1}{c}{\xmark}                                                          & \multicolumn{1}{c}{\xmark}                                               & \multicolumn{1}{c}{\cmark} \\ \hline
\textit{VFedTrans}     & \multicolumn{1}{c}{\cmark}                                                     & \multicolumn{1}{c}{\cmark}                                                          & \multicolumn{1}{c}{\xmark}                                               & \multicolumn{1}{c}{\xmark} \\ \hline
\textit{APC-VFL}       & \multicolumn{1}{c}{\cmark}                                                     & \multicolumn{1}{c}{\cmark}                                                          & \multicolumn{1}{c}{\cmark}                                               & \multicolumn{1}{c}{\cmark} \\ \hline
\end{tabular}
}
\label{tab:vfl data usage and comm}
\end{table}


\subsection{Autoencoders}
Since autoencoders play a fundamental role in our proposal, we consider that having a basic understanding of this architecture is a must. Autoencoders \cite{bank2023autoencoders} are a type of neural network with a wide range of applications, such as representation learning or anomaly detection \cite{berahmand2024autoencoders}.

The architecture consists of two main components: an encoder, $g$, and a decoder, $f$. Given an input vector $\mathbf{x}$, the autoencoder aims to produce an output $\Tilde{\mathbf{x}}$ such that $\mathbf{x} \approx \Tilde{\mathbf{x}}$. The model avoids simply learning an identity function by employing transformations within both the encoder and decoder. Specifically, for an input $\mathbf{x} \in \mathbb{R}^N$, the encoder maps it to a space with a different dimensionality, i.e., $g(\cdot): \mathbb{R}^N \to \mathbb{R}^M$, where $N \neq M$, resulting in a \textit{latent representation} of the input, commonly denoted as $\mathbf{z}$. From this latent space, the decoder reconstructs the input, and can be defined as $f(\cdot): \mathbb{R}^M \to \mathbb{R}^N$ \cite{michelucci2022introduction}. Based on this, the learning objective of the autoencoder can be formalized:

\begin{equation}
    \operatorname*{arg\,min}_{f,g} \frac{1}{|\mathcal{D}|} \sum_{i=1}^{|\mathcal{D}|} \Delta(\mathbf{x}_i, f(g(\mathbf{x}_i))), \quad i = 1, 2, \ldots, |\mathcal{D}|
\end{equation}

where,
\begin{description}
    \item[$\Delta(\cdot)$] is a function that quantifies the difference between the input and the output, i.e., Mean Squared Error (MSE).
    \item[$|\mathcal{D}|$] is the number of samples used for training.
\end{description}

Given that the latent representation is $\mathbf{z} \in \mathbb{R}^M$, based on its dimension, the autoencoder can be either undercomplete or overcomplete. If the dimension of the latent space is greater than the input dimension ($M > N$), the autoencoder is defined as an overcomplete autoencoder, otherwise, it is an undercomplete autoencoder \cite{charte2018practical}.

\section{Problem definition}
Consider an scenario where $K$ participants collaboratively train a classification model using a vertically partitioned dataset. For the $k$-th participant, let $\mathcal{X}_k$ denote the feature space, $\mathcal{Y}_k$ the label space, and $\mathcal{I}_k$ the ID space, such that its dataset is represented as a triplet $\mathcal{D}_k = \{X_k, \mathbf{y}_k, \mathbf{i}_k\}$, where $X_k \in \mathcal{X}_k,\;\mathbf{y}_k\in\mathcal{Y}_k\;\text{and}\ \mathbf{i}_k\in\mathcal{I}_k$.

In VFL, a single participant, $k=K$, known as active participant, is the only one with access to labels. The rest, $k = 1, 2, ..., K-1$, are passive participants and only have features and IDs. This allows to redefine the datasets, $\mathcal{D}_k$: 

\begin{equation}
    \mathcal{D}_k = 
\begin{cases} 
\{X_k, \mathbf{y}_k, \mathbf{i}_k\} & \text{if } k = K, \\
\{X_k, \mathbf{i}_k\} & \text{otherwise}. \\
\end{cases}
\end{equation}

A defining characteristic of VFL is that the feature spaces of the participants must be different while the ID space must be shared, allowing participants to align their respective samples:
\begin{equation}
\mathcal{X}_i \ne \mathcal{X}_j, \mathcal{I}_i = \mathcal{I}_j \forall i\ne j    
\end{equation}

An assumption that is done in classical VFL is that all participants share the same set of sample IDs, $\mathbf{i}_i=\mathbf{i}_j\forall i\ne j$. This implies that every sample exists across all datasets. However, this is a strong and often unrealistic assumption in many real-world scenarios.

We assume that the participants conduct a private set intersection (PSI) process based on the IDs \cite{morales2023private}, which allows each one of them to divide their local dataset in two disjoint sets, $\mathcal{D}_{k_a}$ and $\mathcal{D}_{k_u}$, corresponding to the aligned and unaligned samples respectively. Note that $\mathcal{D}_{k_a} \ne \varnothing\; \forall k$, as having a non-empty set of aligned samples is a must in order to be able to collaborate in a VFL training process. The focus of classical VFL has been the aligned data among participants, 
$\mathcal{D}_A$, as represented in Figure \ref{fig: data VFL}.

\begin{equation}
\mathcal{D}_A = \bigcup_{k = 1}^K \mathcal{D}_{k_a}    
\end{equation}

However, this approach is unable to incorporate the knowledge of the unaligned samples to the federation process nor perform inference on them. The problem being addressed in our work, is overcoming this limitation while maintaining communication efficiency.


\begin{figure*}
  \centering
  \includegraphics[width=.95\textwidth]{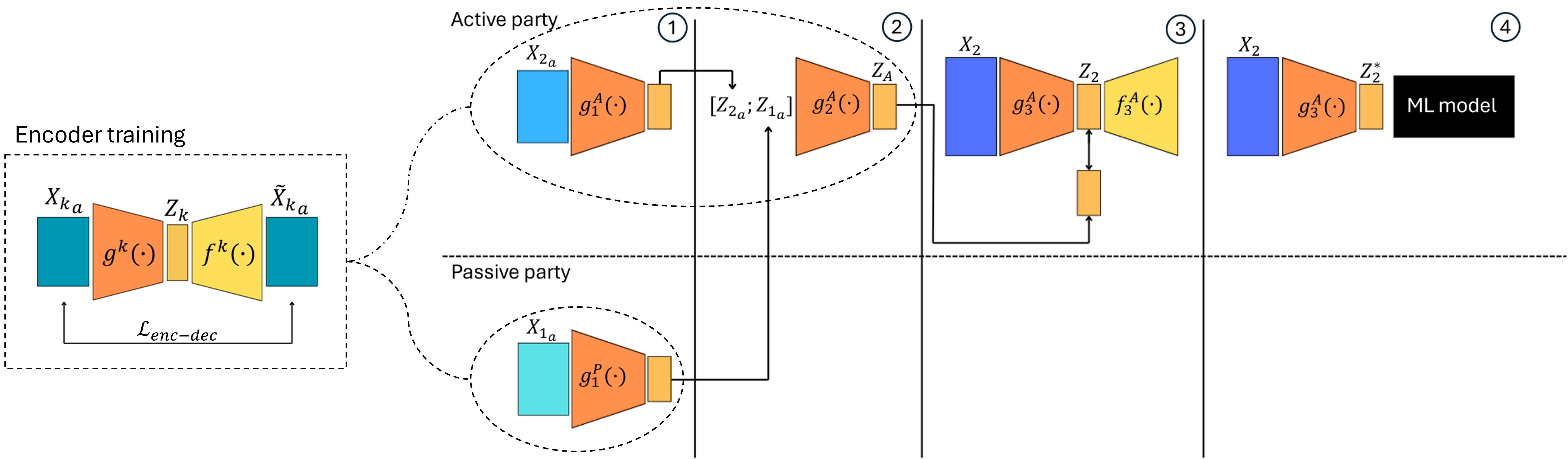}
  \caption{Proposed VFL process with APC-VFL, comprising four main steps. \textcircled{1}: local representation learning. \textcircled{2}: aligned representation learning. \textcircled{3}: knowledge distillation. \textcircled{4}: final classification model training.}
  \label{fig:proposal flow}
\end{figure*}

\section{Method}
\label{sec:method}
We propose a VFL method that combines the usage of latent representations and knowledge distillation. It is designed to minimize the amount of communication rounds among participants, while improving active party's model performance even when a fraction of samples are aligned among participants.  As illustrated in Figure \ref{fig:proposal flow}, the proposed approach consists of four steps, requiring information exchange between participants only in the first one. 


In this section, we describe the general idea behind our proposal and detail each of its steps.

\subsection{General description}
Although the proposal consists of four distinct steps, its primary objective is to learn a joint representation of the aligned samples and transfer this knowledge to the active participant.

To achieve this, each participant independently learns representations of its aligned samples locally, without sharing any information with other participants \textcircled{1}. Once these representations are learned, they are sent to the active participant. This represents the only information exchange in APC-VFL. Using the received representations, the active participant learns a joint representation \textcircled{2}, which then conditions the training of an encoder \textcircled{3}. This encoder generates enhanced representations of the active participant's entire dataset that are subsequently used to train a classification model \textcircled{4}.

This approach enables the use of non-aligned samples during training, as the classifier is trained on the active participant’s complete dataset. Furthermore, because the classifier is trained on the active participant’s enhanced data, produced through the federated learning process, inference can also be performed using the active participant’s data exclusively. This allows the active participant to perform inference in an independent, non collaborative, manner.
\subsection{Local representation learning}


While the choice of architecture is highly application-dependent, it is important to note that tabular datasets, the most common ones in the VFL domain, are typically low-dimensional compared to other data modalities, such as images. Within tabular datasets, federated learning is especially relevant in scenarios where participants have limited features, as it can report greater improvements with respect to locally trained models. In these cases, further reducing the dimensionality of the dataset is impractical, making overcomplete autoencoders the preferred choice for the representation learning process.

With a trained autoencoder, the local representations of the aligned samples can be obtained by performing a forward pass through the encoder: $Z_{K_a} = g_1^K(X_{k_a})$, being $g_1^k$ the trained encoder of the $k^{th}$ participant, as depicted in Fig \ref{fig:proposal flow}.






\subsection{Aligned representation learning and knowledge distillation}

Even if these are two different steps of the proposal, they are highly correlated and therefore, it is interesting to analyze them as one.

Firstly, the joint representations of the aligned samples have to be learned by means of $g_2^K$, which once again is proposed to be trained using the autoencoder architecture, as in the first step. Given that the input of this encoder is in a space of dimension $\mathbb{R}^{N_2}$, being $N_2$ the sum of all the latent dimensions of the local encoders, it is reasonable to use $g_2^K$ to reduce the dimension of the input, learning compact, mixed, joint representations. 


Having the second encoder trained, for the knowledge distillation process it must be taken into account that $g_2^K(\cdot): \mathbb{R}^{N_2} \to \mathbb{R}^{M_2}$. Similarly, the student encoder, which is the final encoder is a function $g_3^K(\cdot): \mathbb{R}^{N_3} \to \mathbb{R}^{M_3}$, being $\mathbb{R}^{N_3}$, the dimensionality of the active participant´s features. To align these representations for feature-based knowledge distillation \cite{Gou_2021}, we set $M_3=M_2$. The dimensional consistency avoids the need for additional regressors and ensures that the teacher and student models produce compatible representations.

In this case, the distillation is not intended to transfer the knowledge from a model with high capacity to a small one, but rather to transfer what could be considered privileged knowledge, the one of the joint latent representation, to a model that has no access to that data. In order to carry this distillation process out, a comparison between latent representations generated by $g_2^K$ and $g_3^K$ is done, applying the MAE error as proposed in \cite{Huang_2023} or MSE error, which is formalized in Equation \ref{equ: kd} for the $i^{th}$ data point of the active party, $\mathbf{x}_i$.

\begin{equation}
\label{equ: kd}
    \resizebox{0.99\hsize}{!}{%
    $
    \mathcal{L}_{\text{total}}(\mathbf{x}_i) = 
     \begin{cases}
       \mathcal{L}_{\text{enc-dec}}(\mathbf{x}_i) + \lambda \cdot \mathcal{L}_{\text{distill}}(\mathbf{x}_i), & \text{if } \mathbf{x}_i \text{ is aligned} \\
       \mathcal{L}_{\text{enc-dec}}(\mathbf{x}_i), & \text{otherwise}
     \end{cases}
     $
     }
\end{equation}

where,
\begin{description}
    \item[$\mathcal{L}_{enc-dec}(\mathbf{x}_i) = \lvert \lvert \mathbf{x}_i - f_3^K(g_3^K(\mathbf{x}_i)) \rvert\rvert_2^2$] 
    \item[$\mathcal{L}_{distill}(\mathbf{x}_i) =  
     \begin{cases}
       \lvert \lvert \mathbf{z}_{A_i} - g_3^K(\mathbf{x}_{i}) \rvert \rvert_2^2  &\hspace{-1mm}\text{\normalfont if MSE as loss}\\
       \lvert \mathbf{z}_{A_i} - g_3^K(\mathbf{x}_{i}) \rvert  &\hspace{-1mm}\text{\normalfont if MAE as loss}\\
     \end{cases}$]
    \item [$\lambda \in \mathbb{R}^+$ \normalfont is the weight of the distillation loss term]
\end{description}

Notice how the loss to train the final encoder with the autoencoder architecture differs for those data points that are aligned and for those that are not. This is because the joint latent representation is known if and only if the data point is aligned, i.e. $\exists\mathbf{z}_{A_i}\iff\mathbf{x}_i\in\mathcal{D}_A$, and therefore, a comparison among the representations can only be done in that scenario, limiting the loss to the reconstruction one, $\mathcal{L}_{enc-dec}$, for the unaligned samples.

By having a composed loss function, $\mathcal{L}_{total}$, the autoencoder cannot only learn how to reconstruct the input. Conditioning the latent representations with $\mathcal{L}_{distill}$ from Equation \ref{equ: kd}, the autoencoder, and more precisely the encoder, needs to learn to generate \textit{enriched} data representations. These representations are considered to be \textit{enriched}, since they have to resemble to the joint aligned representations having as input just the information of the active party.


\subsection{Classification model training}
After training the final autoencoder, the active participant only needs to generate the enhanced dataset required to train the final classifier. To do this, the participant obtains the representations, $Z_K$, of all the samples' features, $X_K$, using the trained encoder $g_3^K$: $Z_K=g_3^K(X_K)$.

With these representations, the enhanced dataset can be constructed as follows: $\mathcal{D}_K^*=\{Z_K, \mathbf{y}_k, \mathbf{i}_k\}$.  Since the identifiers lack predictive capability, the final classifier can be trained like any other machine learning or deep learning model, utilizing the data representations $Z_K$ instead of the original features $X_K$, and known labels, $\mathbf{y}_K$.

Notice how this process can be done autonomously by the active participant, i.e. it does not require any information from the collaborators, which makes the active participant independent.



\subsection{Privacy analysis}

Our work focuses on the algorithmic side of VFL and we assume that participants are \textit{honest but curious}. Under that assumption, it must be taken into account that APC-VFL works in the representation space, and does not require participants to share their original data, local models nor gradients, which can cause information leakage \cite{geiping2020inverting, zhu2019deep}. 

In our proposal, each participant learns a function $g(\cdot): X \to Z$, i.e. a function that maps the data from the feature space to the representations space. Given that there are infinitely many $g(\cdot)$ functions, it is safe to share data representations, $Z$, as long as the function is kept local. Our work keeps the encoders of each participant local, which is equivalent to not sharing the functions and therefore, under the assumption of \textit{honest but curious} participants, the proposal can be deemed to be safe \cite{gupta2018distributed, shankar2024silofusecrosssilosyntheticdata}.

\section{Experimental set-up}
Evaluating federated learning proposals generally, and more precisely the ones of scenarios with vertically partitioned dataset is not trivial, and there is no agreement about how to perform such task \cite{chai2024survey}. This section aims to clearly define our experimental set-up.

\paragraph{Datasets}
We have chosen three tabular datasets, as we consider that tabular data is the most relevant one for the VFL setting. Those datasets are \textit{MIMIC-III}, that has been preprocessed as in \cite{Huang_2023}, which results in 58976 samples with 15 features and 4 target classes \cite{MIMIC}. The second one is is \textit{Breast Cancer Wisconsin}, which has 569 samples with 30 features and 2 target classes \cite{cancer}. The final one is \textit{UCI credit card}, which has a total of 30000 samples with 23 features and binary labels \cite{default_of_credit_card_clients_350}.

\paragraph{Data partitions}
As the chosen datasets are not \textit{VFL native}, they have to be adapted to such scenario. For that, a federated process with two participants, a common set-up for VFL experiments \cite{sun2023communication, liu2020federated}, has been chosen.

Under this set-up, we have generated different scenarios by varying the number of aligned samples between participants. Furthermore, as in \cite{Huang_2023}, when partitioning the datasets we assume that the active party has less available features, which would motivate it to perform a federation process. Given that in our proposal the amount of features that each participant has does make a difference, as the alternative dataset is built based on them, different partitions in terms of features have been also tested for each scenario.

Summarizing, on each one of the datasets, we have defined four different scenarios in terms of sample alignment by varying $|\mathcal{D}_A|$. Within each one of those scenarios, we have modified the features the active participant has access to, i.e. the dimensionality of $X_K$. This has resulted in a total of 48 distinct VFL scenarios, since four different dimensionalities have been tested. Further details about the data partitions can be found on Appendix \ref{app: data partitions}.

\paragraph{Models}
All autoencoders considered in APC-VFL are symmetric, i.e. the encoder-decoder architectures are the same in an inverse order. A logistic regression model has been used as the classifier, which can be helpful to assess the quality of the proposal as is done in \textit{linear probing} \cite{alain2018understandingintermediatelayersusing}. A detailed description of the architectures and hyperparameters can be found on Appendix \ref{app: hyperparameters}.

\section{Evaluation}
The performed experiments can be divided into two main groups: comparisons with VFedTrans and locally trained models with limited sample alignment; and comparison with Vertical SplitNN.

While the comparison with VFedTrans and local models is a must, comparing our proposal with Vertical SplitNN may seem odd. The rationale for comparing our approach with a classical VFL method, such as Vertical SplitNN, is to evaluate its performance under a traditional VFL setting, where it is assumed that data is fully aligned.

\begin{figure}[!h]
    \centering
    \includegraphics[width=1\columnwidth]{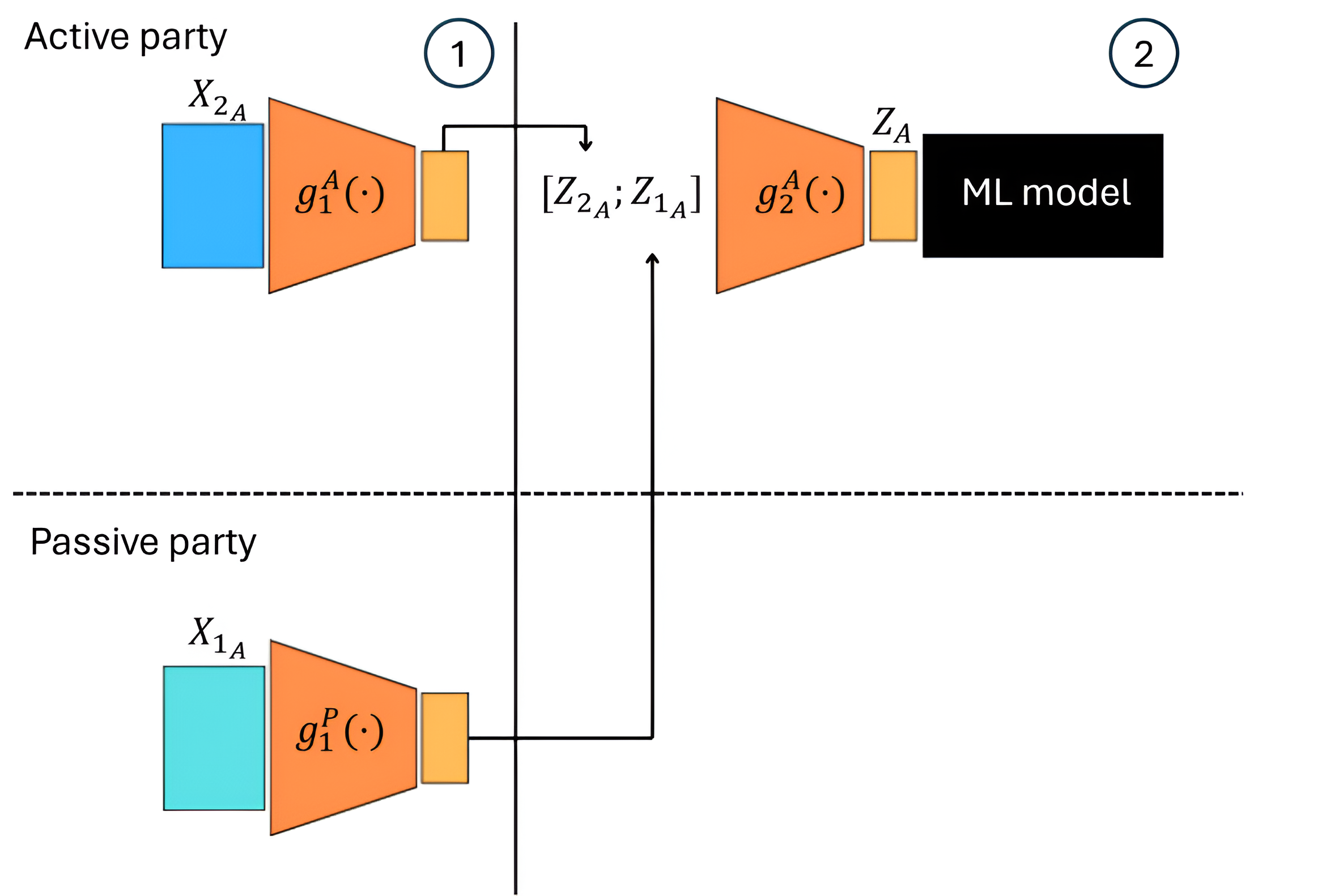}
    \vspace{-7mm}
    \caption{Adjustment of the proposal for the comparison with vertical splitNN and assess the quality of the joint latent representations.}
    \label{fig: proposal comp splitNN}
\end{figure}

Having every sample aligned, we have made slight adjustments to our approach for this scenario by training the classification model on the joint latent representations, as illustrated in Figure \ref{fig: proposal comp splitNN}. Given that there are no unaligned samples, the knowledge distillation process would not be able to increase the performance of the final encoder used to learn enhanced data representations. Therefore, it has been omitted and the final classification model has been trained on the joint data representations, which allows to evaluate the usefulness of local representation learning.

    

\paragraph{Evaluation process}
To evaluate the effectiveness of our proposed method in comparison with locally trained models and VFedTrans, we performed a 10-fold cross-validation on the active participant's local data to establish baseline performance for the locally trained models.

\begin{figure*}[ht]
  \centering
  \includegraphics[width=1\textwidth]{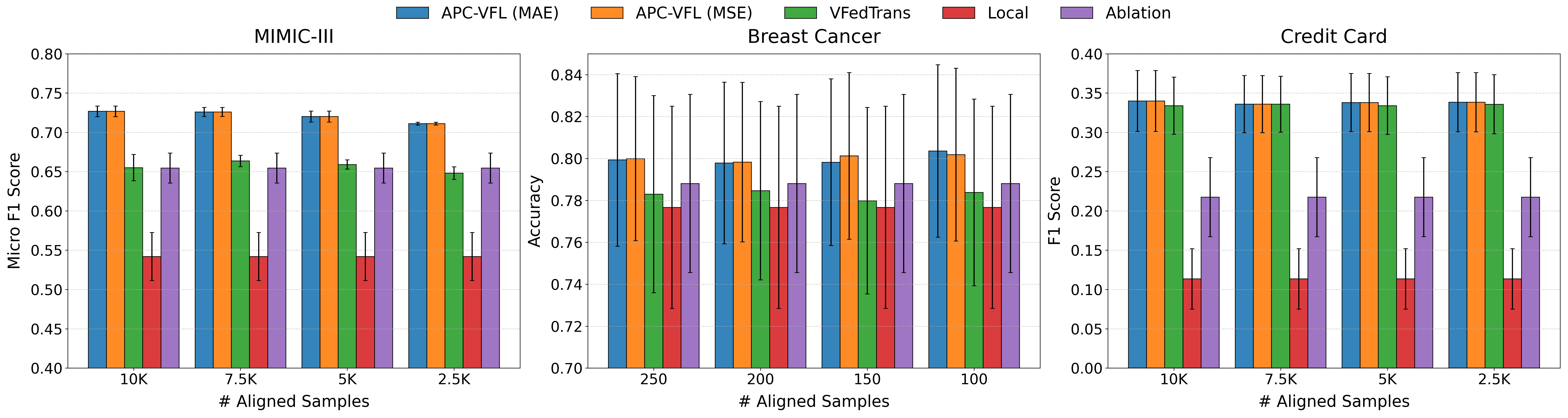}
  \vspace{-6mm}
  \caption{Mean results of the tested data partitions with different quantities of aligned samples. In APC-VFL, the loss used for knowledge distillation is shown between parenthesis. \textit{Macro} and \textit{weighted} averaged results for MIMIC-III can be found on Appendix \ref{app: extra results}.}
  \label{fig: results federated present}
\end{figure*}


When comparing APC-VFL and VFedTrans, we conducted the same 10-fold cross-validation, but on the enhanced dataset, which consists of the representations obtained from the final encoder, $g_3^A$. Since this encoder applies a non-linear transformation to the data, potentially improving the performance of the final classifier, we conducted an ablation study to isolate the impact of the distillation process and the overall federated training, which results are presented as \textit{Ablation} on the plots. In this study, $g_3^A$ was trained without the distillation loss, using only the local data.
 
All experiments were repeated five times for consistency, i.e. five training processes with 10-fold cross-validation were performed. The final results are reported as the mean of the means across the five runs, $\mu_{result} =\frac{1}{5} \sum_{i = 1}^5\mu_i$, along with their standard deviation.

The comparison with vertical splitNN has been done in a different way. The datasets have been reduced to the aligned samples, as that is the assumption done in classical VFL. With those datasets, the training and testing process has been done 5 times, of which the mean and standard deviation of the evaluation metrics as well as communication round amount have been computed. Given that for this scenario the number of features that each participant has does not affect, since the training and inference process is collaborative, we have simply varied the number of aligned samples, which is equivalent to varying the dataset size on this case.

\subsection{Comparison with Local training and VFedTrans}

The main results for this comparison are in Figure \ref{fig: results federated present}. Notice how the previously mentioned 48 scenarios have been reduced to 16, by computing the mean metric of the different vertical partitions, i.e the mean metric for each $|\mathcal{D}_A|$.

\begin{figure}[h]
  \centering
  \includegraphics[width=0.47\textwidth]{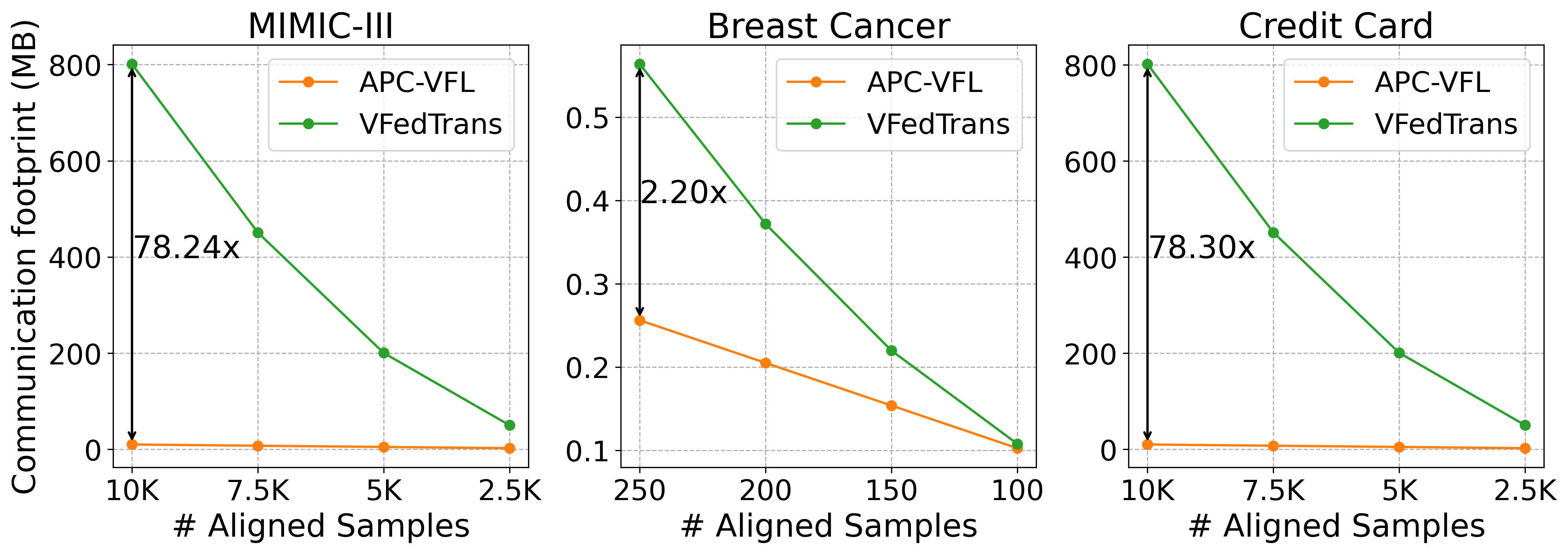}
  \caption{Communication footprint of APC-VFL vs VFedTrans. The proposed APC-VFL achieves large communication savings while performing at least as well as VFedTrans as Fig \ref{fig: results federated present} shows.} 
  \label{fig:comm mimic}
\end{figure}

APC-VFL, consistently outperforms not only the locally trained models, as expected, but VFedTrans as well in MIMIC-III and Breast Cancer. Moreover, APC-VFL achieves this performance advantage over VFedTrans while significantly reducing the communication overhead of the training process across all datasets, by up to $78.30\times$, as illustrated in Figure \ref{fig:comm mimic}.

While VFedTrans requires a reduced and constant number of communication rounds for the federated representation learning process, its communication overhead increases exponentially with the number of aligned samples. In contrast, our approach requires only a single communication round, and its communication footprint scales linearly with the number of aligned samples.

\begin{figure}[h]
  \centering
  \includegraphics[width=0.48\textwidth]{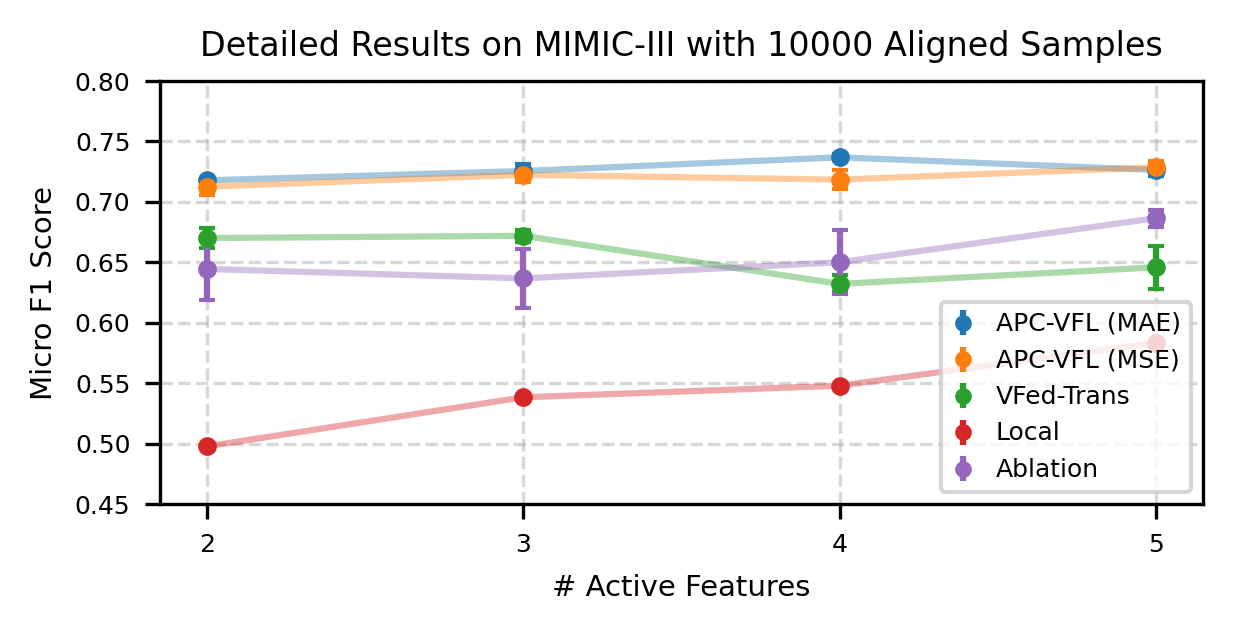}
  \vspace{-7mm}
  \caption{Detailed plot of MIMIC-III with 10000 aligned samples.}
  \label{fig:fine-grained MIMIC 10K}
\end{figure}

To gain deeper insights from these general plots, it is important to understand the meaning of the standard deviation shown on the bars: it reflects the change in model performance when the number of features available to the active participant varies. Smaller deviations indicate greater invariance across data partitions, with this effect being particularly noticeable in the MIMIC-III dataset. In these cases, the distillation process in our approach effectively compensates for the lack of available features. This can be visually observed in Figure \ref{fig:fine-grained MIMIC 10K}, which presents a fine-grained view of the scenario with 10000 aligned samples. Additional fine-grained representations of the results are provided in Appendix \ref{app: fine-grained plots}.

Figure \ref{fig:fine-grained MIMIC 10K} also allows us to confirm that as previously stated, as the active participant has less available features, performing a federated learning process is more valuable, since the difference between the local model and the federated one is greater. On the other hand, this fine-grained plot also allows us to identify one of the main limitations of VFedTrans: the embedding dimension constraint. The embedding dimension, which is known to play a crucial role in representation learning \cite{sui2024selfsupervised}, is defined by the dimensionality of the federated dataset for VFedTrans, unlike in our approach, which allows to set the embedding dimension based on the autoencoder architecture. This inherent limitation of VFedTrans is the reason why the \textit{Ablation} curve, which is the result of an enhanced local model, is able to outperform it on some settings.

\begin{figure}[h]
  \centering
  \includegraphics[width=0.48\textwidth]{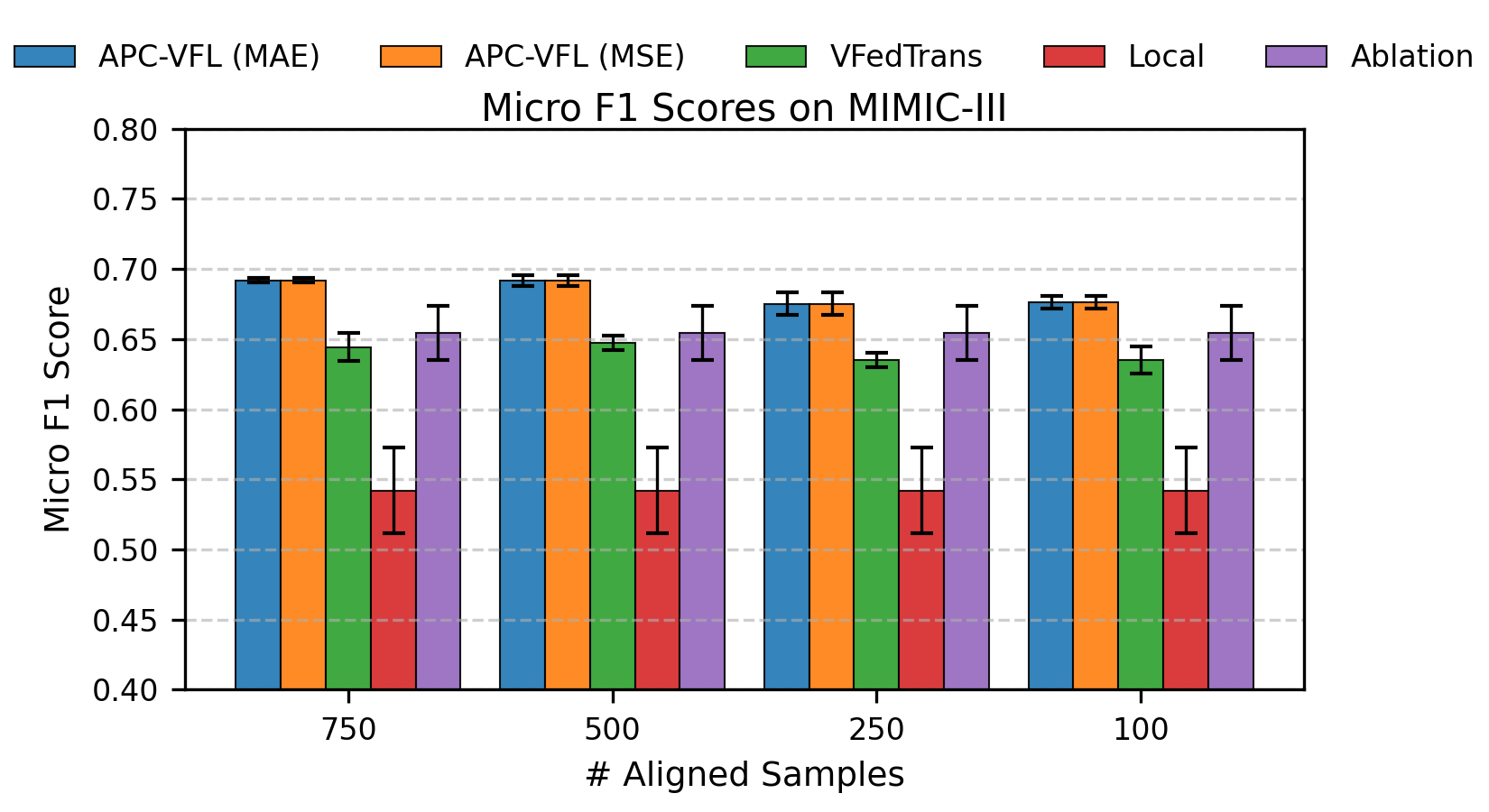}
  \vspace{-6mm}
  \caption{Results on MIMIC-III with reduced aligned samples demonstrating APC-VFL superiority. It is an extension to Fig.~\ref{fig: results federated present}.}
  \label{fig:mimic micro F1 minimum overlap}
\end{figure}

Lastly, it must also be noted that while the locally trained models, represented by \textit{Local} and \textit{Ablation} bars, are not affected by the number of aligned samples, the distillation based approaches also suffer little performance decay as this number is reduced, being this decay the greatest for the dataset MIMIC-III (see Figure \ref{fig: results federated present}). This has motivated us to define a set of reduced aligned samples $\{750, 500, 250, 100\}$ and repeat the experiments on this dataset. The mean results are reported on Figure \ref{fig:mimic micro F1 minimum overlap}.


It can be observed that on these scenarios VFedTrans’ performance is slightly worse then the enhanced local model's. On the other hand, APC-VFL still shows improvements with respect to local models, even with as few as 100 aligned samples, which represent the $0.5\%$ of the total samples.

   
   

\subsection{Results of the comparison with vertical splitNN}
When analyzing these results, it must be taken into account that they have been obtained on an ideal, classical scenario: there were no \textit{strugglers} nor loss of information during the training process, which would degrade the performance of Vertical SplitNN.

\begin{table}[!h]
\tiny
\centering
\caption{Results of the comparison between SplitNN and APC-VFL with different amounts of aligned samples across datasets. Results are shown as Mean $\pm$ Std.}
\vspace{3mm}
\label{tab:merged_results}
\scalebox{1.095}{
\begin{tabular}{l l c c c }
    \toprule
    \textbf{\# Alig.} & \textbf{Method} & \textbf{Metric} & \textbf{\# Comm} & \textbf{Cost (MB)} \\
    \midrule
    \multicolumn{5}{c}{MIMIC-III (Micro F1-score)} \\
    \midrule
    \multirow{2}{*}{10K} 
        & SplitNN & $0.826 \scalebox{0.9}{\(\pm 0.002\)}$ & $4290 \scalebox{0.9}{\(\pm 640\)}$ & $561.57 \scalebox{0.9}{\(\pm 83.86\)}$\\ 
        & APC-VFL & $0.745 \scalebox{0.9}{\(\pm 0.003\)}$ & $1$ & 9.73\\ 
    \midrule
    \multirow{2}{*}{7.5K} 
        & SplitNN & $0.770 \scalebox{0.9}{\(\pm 0.037\)}$ & $3146 \scalebox{0.9}{\(\pm 470\)}$ & $381.04 \scalebox{0.9}{\(\pm 198.87\)}$\\ 
        & APC-VFL & $0.747 \scalebox{0.9}{\(\pm 0.004\)}$ & $1$ & 7.17\\ 
    \midrule
    \multirow{2}{*}{5K} 
        & SplitNN & $0.723 \scalebox{0.9}{\(\pm 0.004\)}$ & $634 \scalebox{0.9}{\(\pm 124\)}$ & $82.40 \scalebox{0.9}{\(\pm 16.11\)}$\\ 
        & APC-VFL & $0.742 \scalebox{0.9}{\(\pm 0.009\)}$ & $1$ & 4.61\\ 
    \midrule
    \multirow{2}{*}{2.5K} 
        & SplitNN & $0.698 \scalebox{0.9}{\(\pm 0.006\)}$ & $563 \scalebox{0.9}{\(\pm 143\)}$ & $73.24 \scalebox{0.9}{\(\pm 18.56\)}$\\ 
        & APC-VFL & $0.728 \scalebox{0.9}{\(\pm 0.005\)}$ & $1$ & 2.05\\ 
    \midrule
    \multicolumn{5}{c}{Breast Cancer Wisconsin (Accuracy)} \\
    \midrule
    \multirow{2}{*}{250} 
        & SplitNN & $1.000 \scalebox{0.9}{\(\pm 0.000\)}$ & $380 \scalebox{0.9}{\(\pm 60\)}$ & $27.04 \scalebox{0.9}{\(\pm 4.27\)}$\\ 
        & APC-VFL & $1.000 \scalebox{0.9}{\(\pm 0.000\)}$ & $1$ & $0.26$\\ 
    \midrule
    \multirow{2}{*}{200} 
        & SplitNN & $0.980 \scalebox{0.9}{\(\pm 0.000\)}$ & $312 \scalebox{0.9}{\(\pm 61\)}$ & $22.26 \scalebox{0.9}{\(\pm 4.34\)}$\\ 
        & APC-VFL & $0.976 \scalebox{0.9}{\(\pm 0.008\)}$ & $1$ & $0.20$\\ 
    \midrule
    \multirow{2}{*}{150} 
        & SplitNN & $0.952 \scalebox{0.9}{\(\pm 0.010\)}$ & $156 \scalebox{0.9}{\(\pm 0\)}$ & $11.22 \scalebox{0.9}{\(\pm 0.00\)}$\\ 
        & APC-VFL & $0.964 \scalebox{0.9}{\(\pm 0.015\)}$ & $1$ & $0.15$\\ 
    \midrule
    \multirow{2}{*}{100} 
        & SplitNN & $0.912 \scalebox{0.9}{\(\pm 0.010\)}$ & $84 \scalebox{0.9}{\(\pm 0\)}$ & $6.16 \scalebox{0.9}{\(\pm 0.00\)}$\\ 
        & APC-VFL & $0.956 \scalebox{0.9}{\(\pm 0.015\)}$ & $1$ & $0.10$\\ 
    \midrule
    \multicolumn{5}{c}{Credit Card (F1-score)} \\
    \midrule
    \multirow{2}{*}{10K} 
        & SplitNN & $0.389 \scalebox{0.9}{\(\pm 0.009\)}$ & $1590 \scalebox{0.9}{\(\pm 558\)}$ & $208.13 \scalebox{0.9}{\(\pm 73.05\)}$\\ 
        & APC-VFL & $0.412 \scalebox{0.9}{\(\pm 0.018\)}$ & $1$ & $9.73$\\ 
    \midrule
    \multirow{2}{*}{7.5K} 
        & SplitNN & $0.407 \scalebox{0.9}{\(\pm 0.013\)}$ & $902 \scalebox{0.9}{\(\pm 298\)}$ & $121.24 \scalebox{0.9}{\(\pm 19.58\)}$\\ 
        & APC-VFL & $0.322 \scalebox{0.9}{\(\pm 0.016\)}$ & $1$ & $7.17$\\ 
    \midrule
    \multirow{2}{*}{5K} 
        & SplitNN & $0.401 \scalebox{0.9}{\(\pm 0.077\)}$ & $590 \scalebox{0.9}{\(\pm 195\)}$ & $76.78 \scalebox{0.9}{\(\pm 25.40\)}$\\ 
        & APC-VFL & $0.408 \scalebox{0.9}{\(\pm 0.017\)}$ & $1$ & $4.61$\\ 
    \midrule
    \multirow{2}{*}{2.5K} 
        & SplitNN & $0.187 \scalebox{0.9}{\(\pm 0.090\)}$ & $442 \scalebox{0.9}{\(\pm 128\)}$ & $57.43 \scalebox{0.9}{\(\pm 16.73\)}$\\ 
        & APC-VFL & $0.314 \scalebox{0.9}{\(\pm 0.011\)}$ & $1$ & $2.05$\\ 
    \bottomrule
\end{tabular}
}
\end{table}

As can be observed on Table \ref{tab:merged_results}, the communication rounds that are needed to train a federated model based on SplitNN are significantly grater, up to $4200\times$, than the single round that our adapted approach requires.

Furthermore, the local training based approach not only shows to be much more efficient in terms of communication rounds, but is also able to achieve performance that is on-pair with or even greater than the end-to-end training by means of SplitNN on most tested scenarios.

Therefore, based on the results that have been obtained from this set of experiments, it can be said that the usage of autoencoders for local training, minimizing the required communication rounds and thus, the overall communication overhead, is effective under the classical assumption.

\section{Conclusions and future work}

In this paper we present an efficient and flexible vertical federated learning (VFL) approach. Our method requires a single information exchange between passive and the active participants, does not need a third party server during training and is not limited to the usage of aligned data. The results demonstrate that our approach not only outperforms locally trained models, an essential criterion for federated learning, but also surpasses VFedTrans, a comparable state-of-the-art method in performance and communication cost. Furthermore, it achieves performance on par with SplitNN in traditional VFL settings where data points are aligned across participants. Notably, while SplitNN performs well with large datasets, our approach achieves equivalent, and sometimes superior, results in scenarios with limited data availability, while also offering greater communication efficiency and flexibility.

For future work, we plan to explore alternative representation learning techniques and introduce additional layers of complexity to further enhance performance, all while maintaining the efficiency of the federated process. Additionally, we are interested in investigating the applicability of this VFL approach in multi-modal scenarios, which present a promising avenue for VFL \cite{che2023multimodal}. 


\section*{Impact Statement}
This paper presents work whose goal is to advance the field of Machine Learning. There are many potential societal consequences of our work, none which we feel must be specifically highlighted here.

\bibliography{references}
\bibliographystyle{icml2025}
\newpage
\appendix
\onecolumn
\section{Data partitions} \label{app: data partitions}

For the comparison with \textit{MIMIC-III} the following variables have been chosen at random and initially assigned to the active participant: \textit{Age, NumCalouts, NumDiagnosis, NumProcs and NumCPTevents.} In order to simulate scenarios where the active participant has less features, some of those have been chosen at random, eliminating them from the passive party's dataset and transferring to the active one in the following order: \textit{NumCPTevents, NumProcs and Age.} The dataset has been reduced to 20000 samples and experiments have been performed with 10000, 75000, 5000 and 2500 aligned samples. When comparing with Vertical SplitNN, 500 of the aligned samples have composed the test set.

In the case of \textit{Breast Cancer Wisconsin}, the variables that have initially been assigned to the active party are: \textit{Worst compactness, Concave points error, Smoothness error, Mean texture and Worst fractal dimension.} Being in this case the transfer order the following one: \textit{Worst compactness, Concave points error and Smoothness error.} In this case, 500 samples were taken at random for the active participant, and experiments have been performed assuming that 250, 200, 150 and 100 were aligned. 50 of the aligned samples have composed the test dataset in the comparison with Vertical SplitNN, being this one an extreme set-up in terms of data availability.

For the comparison with \textit{UCI Credit Card} the following variables have been chosen at random and assigned to the active participant for the initial partition: \textit{X3, X5, X7, X9 and X11.} In order to simulate scenarios where the active participant has less features, some of those have been chosen at random, eliminating them from the passive party's dataset and transferring to the active one in the following order: \textit{X3, X5 and X7.} The dataset has been reduced to 20000 samples and experiments have been performed with 10000, 75000, 5000 and 2500 aligned samples. When comparing with vertical splitNN, 500 of the aligned samples have composed the test set.

On every dataset, a $10\%$ of the samples has been used as validation for early stopping.

\section{Hyperparameter configuration} \label{app: hyperparameters}

The encoder architectures can be found in Table \ref{tab:architecture_details} (symmetric autoencoders have been used), Adam \cite{kingma2017adam} has been used as optimizer, with the parameters proposed in the original paper and all the experiments have been carried out with $\lambda = 0.01$.

\begin{table}[h]
\centering
\caption{Autoencoder architecture details. a $\in \{2, 3, 4, 5\}$ and p is the dimension of passive participant's data.}
\begin{tabular}{l l c}
    \hline
    \textbf{Encoder} & \textbf{Layers} & \textbf{Activations} \\ 
    \hline
    $g_1^A$ & {[}a, 64, 128{]} & \multirow{4}{*}{Selu} \\ 
    $g_1^P$ & {[}p, 128, 256{]} &  \\ 
    $g_2^A$ & {[}384, 256, 256{]} &  \\ 
    $g_3^A$ & {[}a, 256, 256{]} &  \\ 
    \hline
\end{tabular}
\label{tab:architecture_details}
\end{table}

When comparing our adapted approach with SplitNN, in order to make a fair comparison, $g_1^A$ and $g_1^P$ have been used as local feature extractors. $g_2^A$ has been used a classification head, adding an extra layer that matches the number of classes to be predicted.

On MIMIC-III and UCI Credit Card, a batch size of 128 has been used to train every model, whereas in Breast Cancer Wisconsin, the encoders have been trained with batches of 8 elements. Every architecture has been trained with a limit of 200 epochs and a \textit{early stopping} based on the loss function with a patience of 10 epochs. This early stopping is the reason why communication rounds and cost for SplitNN, shown in Table \ref{tab:merged_results}, are not constant throughout the iterations.

\section{Detailed plots} \label{app: fine-grained plots}
The decomposition of the plots that can be seen in Figure \ref{fig: results federated present} is presented on this Appendix.

\begin{figure}[h]
    \centering
    \subfigure[Detailed plot of MIMIC-III with 7500 aligned samples.]{%
        \includegraphics[width=0.28\textwidth]{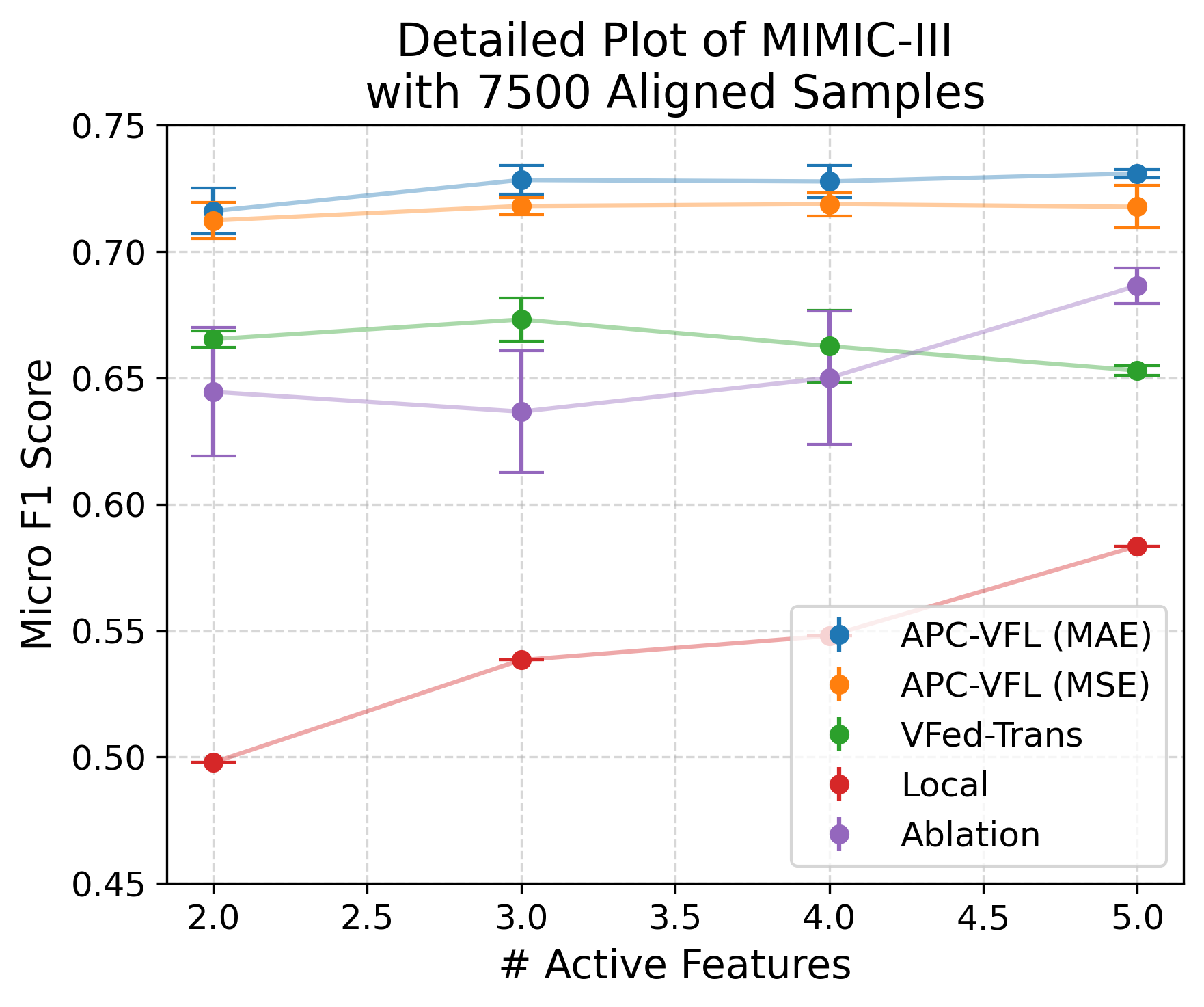}
    }
    \hspace{0.01\textwidth}
    \subfigure[Detailed plot of MIMIC-III with 5000 aligned samples.]{%
        \includegraphics[width=0.28\textwidth]{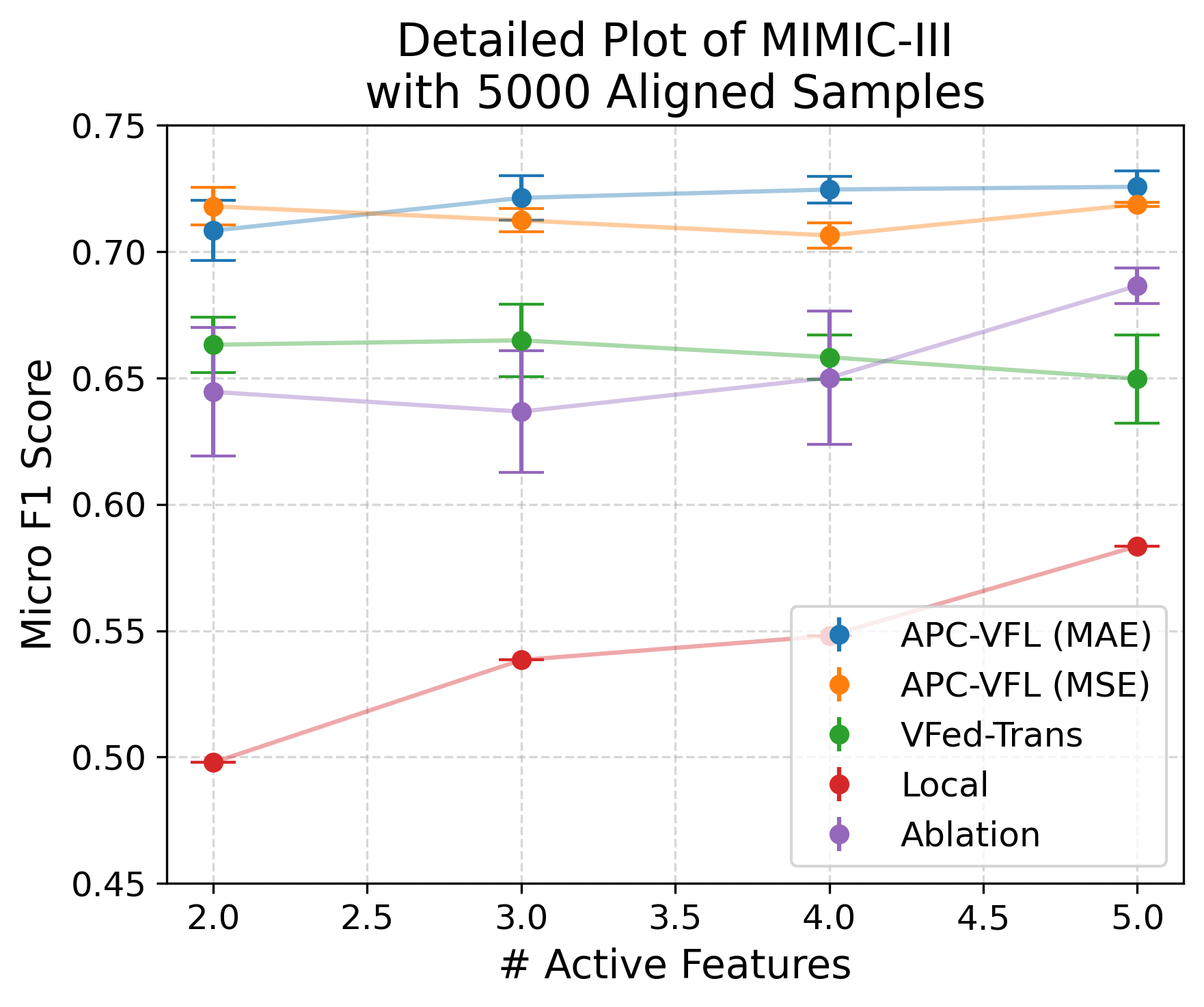}
    }
    \hspace{0.01\textwidth}
    \subfigure[Detailed plot of MIMIC-III with 2500 aligned samples.]{%
        \includegraphics[width=0.28\textwidth]{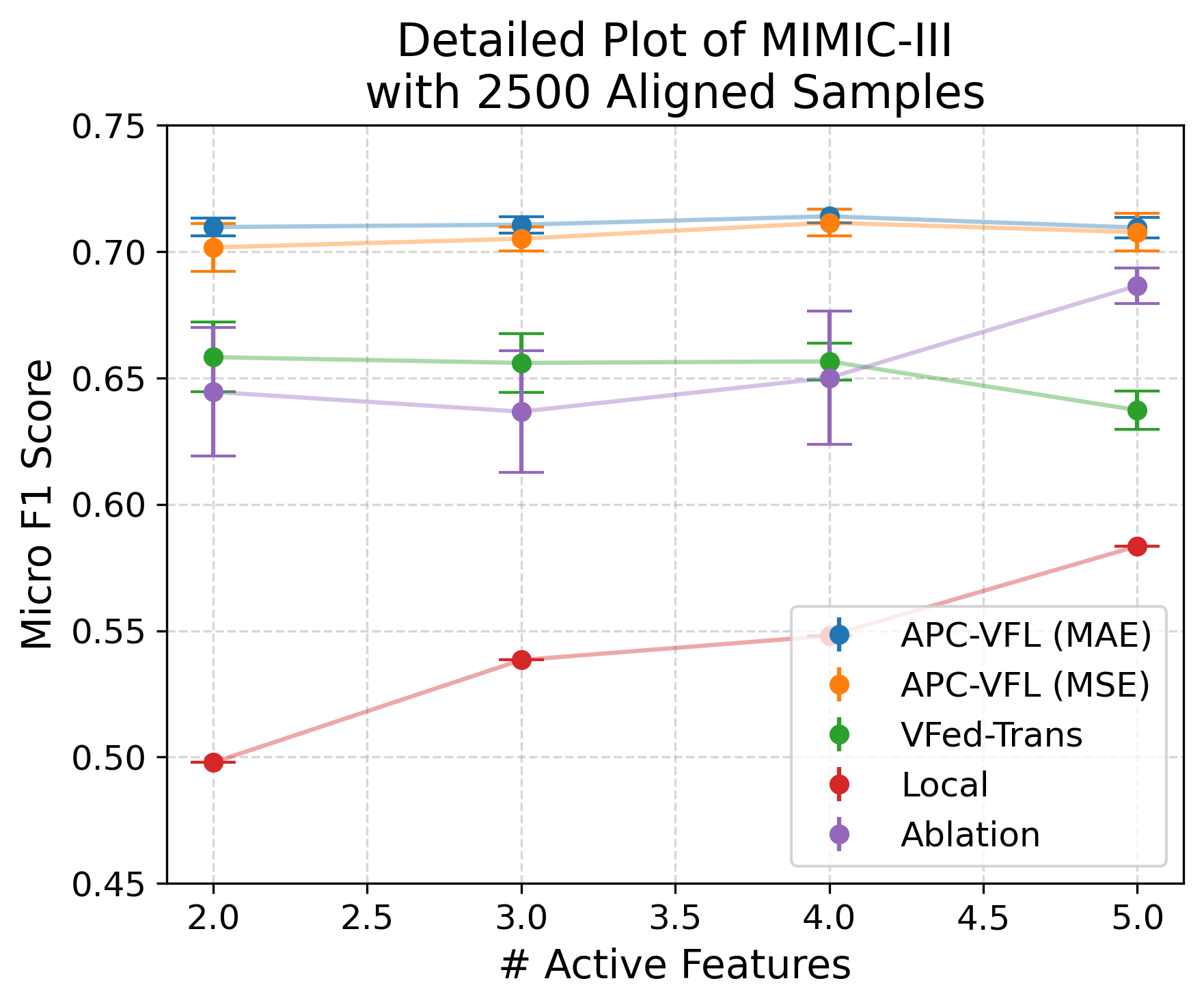}
    }
    \caption{Detailed plots for MIMIC-III with varying sample alignment.}
    \label{fig:fine-graned mimic}
\end{figure}

\begin{figure}[h]
    \centering
    \subfigure[Detailed plot of Breast Cancer Wisconsin with 250 aligned samples.]{%
        \includegraphics[width=0.28\textwidth]{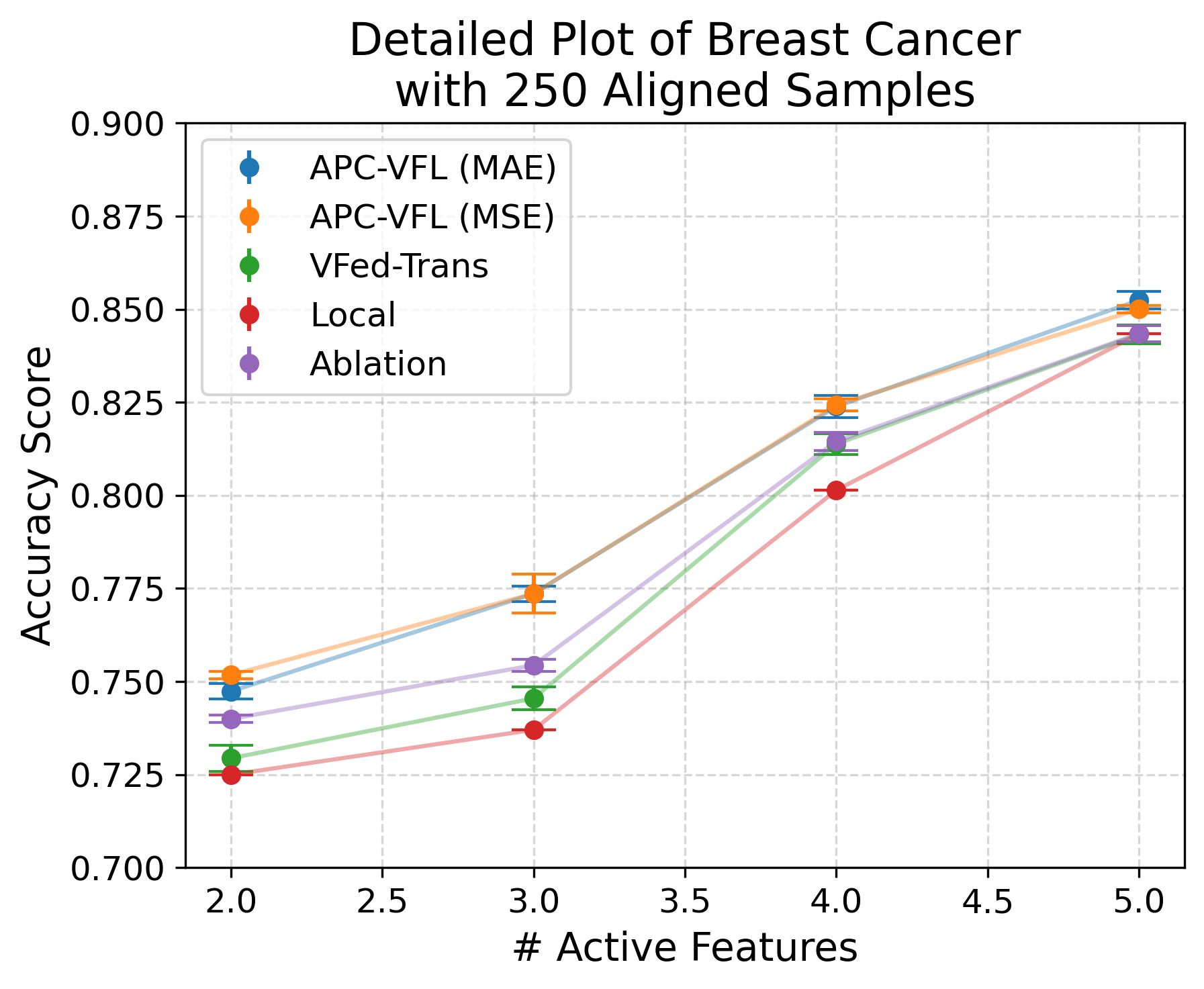}
    }
    \hspace{0.01\textwidth}
    \subfigure[Detailed plot of Breast Cancer Wisconsin with 200 aligned samples.]{%
        \includegraphics[width=0.28\textwidth]{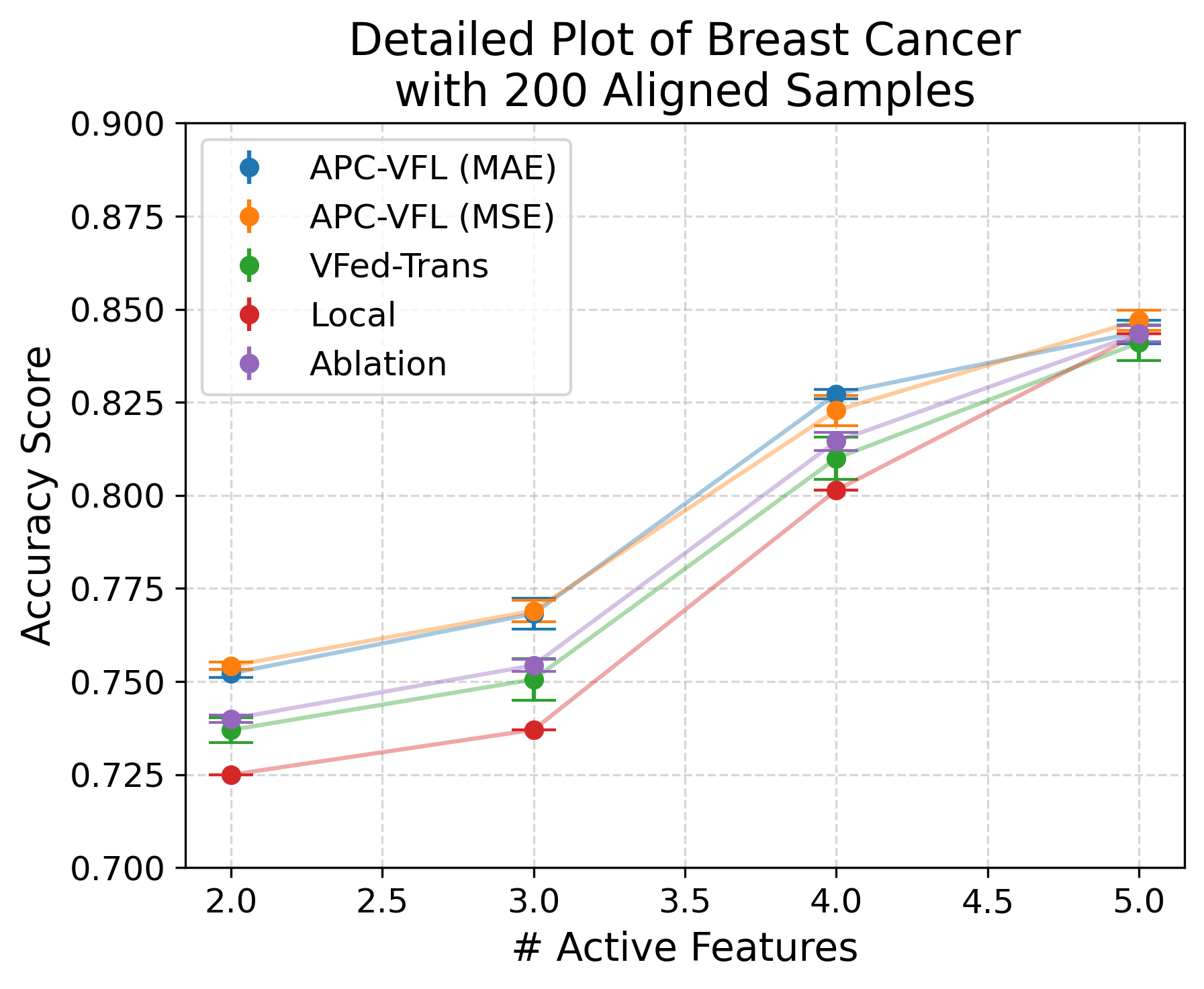}
    }
    \hspace{0.01\textwidth}
    \subfigure[Detailed plot of Breast Cancer Wisconsin with 150 aligned samples.]{%
        \includegraphics[width=0.28\textwidth]{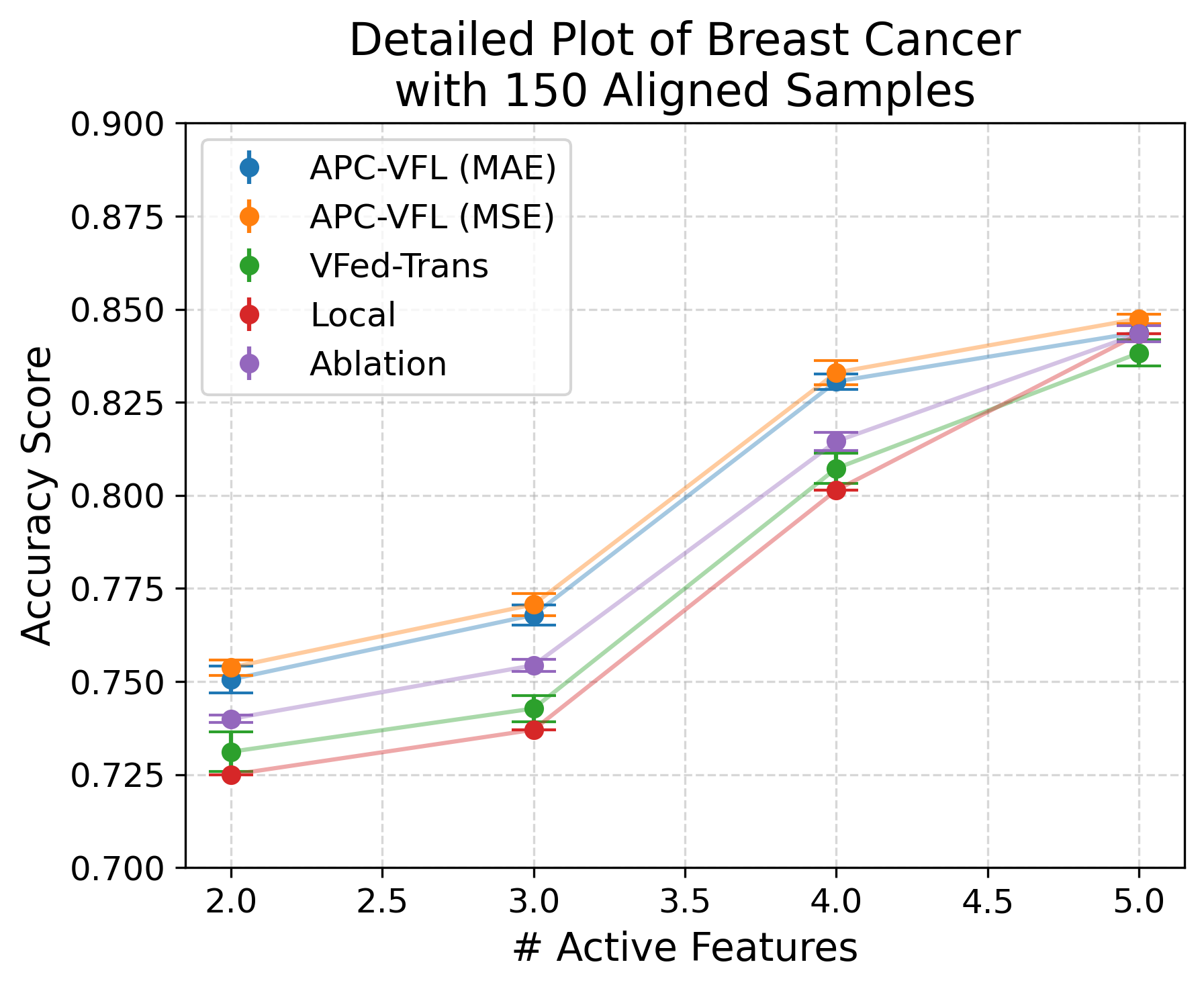}
    }
    \hspace{0.01\textwidth}
    \subfigure[Detailed plot of Breast Cancer Wisconsin with 100 aligned samples.]{%
        \includegraphics[width=0.28\textwidth]{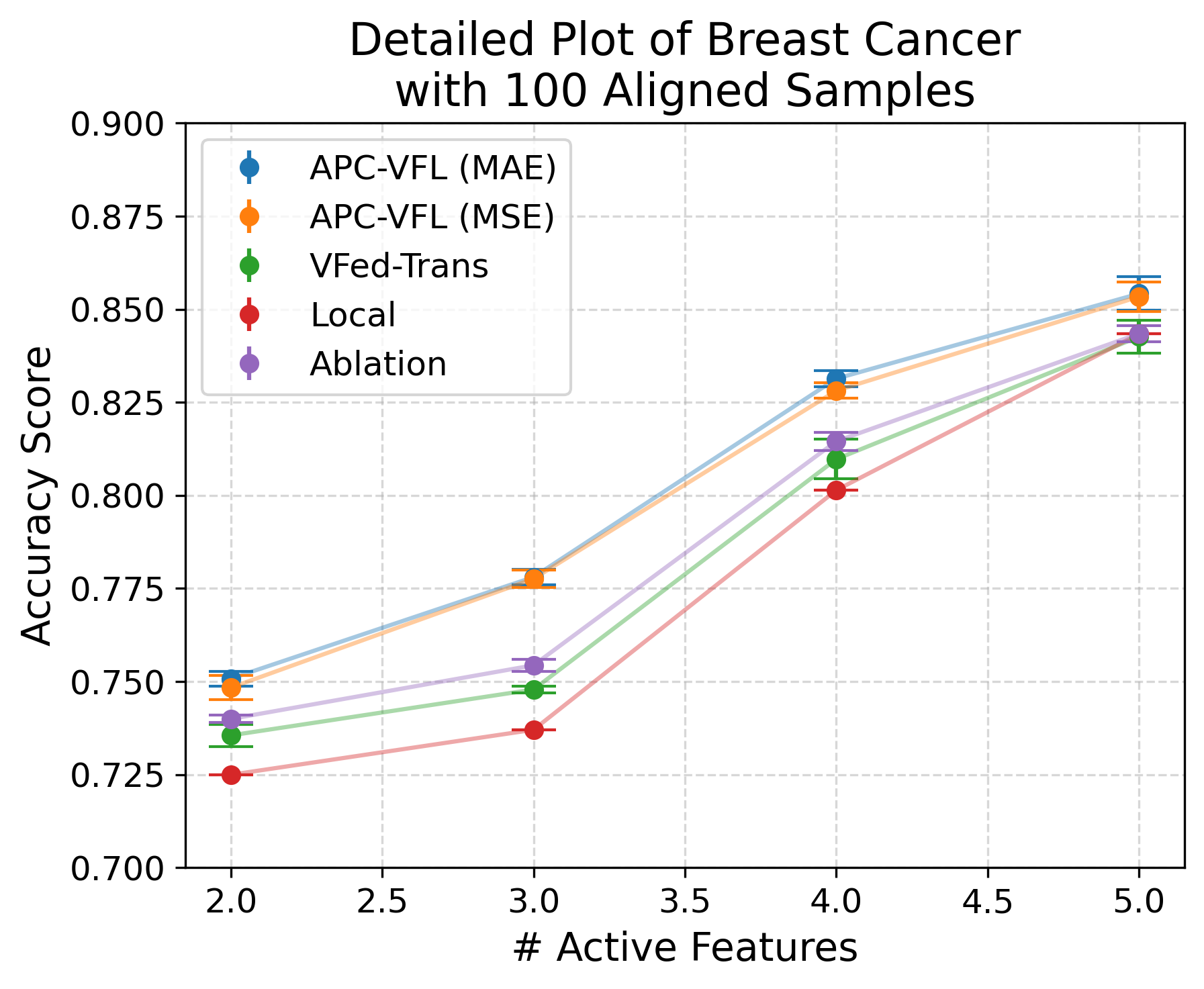}
    }
    \caption{Detailed plots for Breast Cancer Wisconsin with varying sample alignment.}
    \label{fig:fine grained breast cancer}
\end{figure}

Detailed plots for MIMIC-III can be found on Figure \ref{fig:fine-graned mimic}.
\newpage
Figure \ref{fig:fine grained breast cancer} shows the detailed results on Breast Cancer Wisconsin.
\newpage
A detailed representation of the results on UCI Credit Card is shown in Figure \ref{fig:fine grained credit}:
\begin{figure}[h]
    \centering
    \subfigure[Detailed plot of UCI Credit Card with 2500 aligned samples.]{%
        \includegraphics[width=0.28\textwidth]{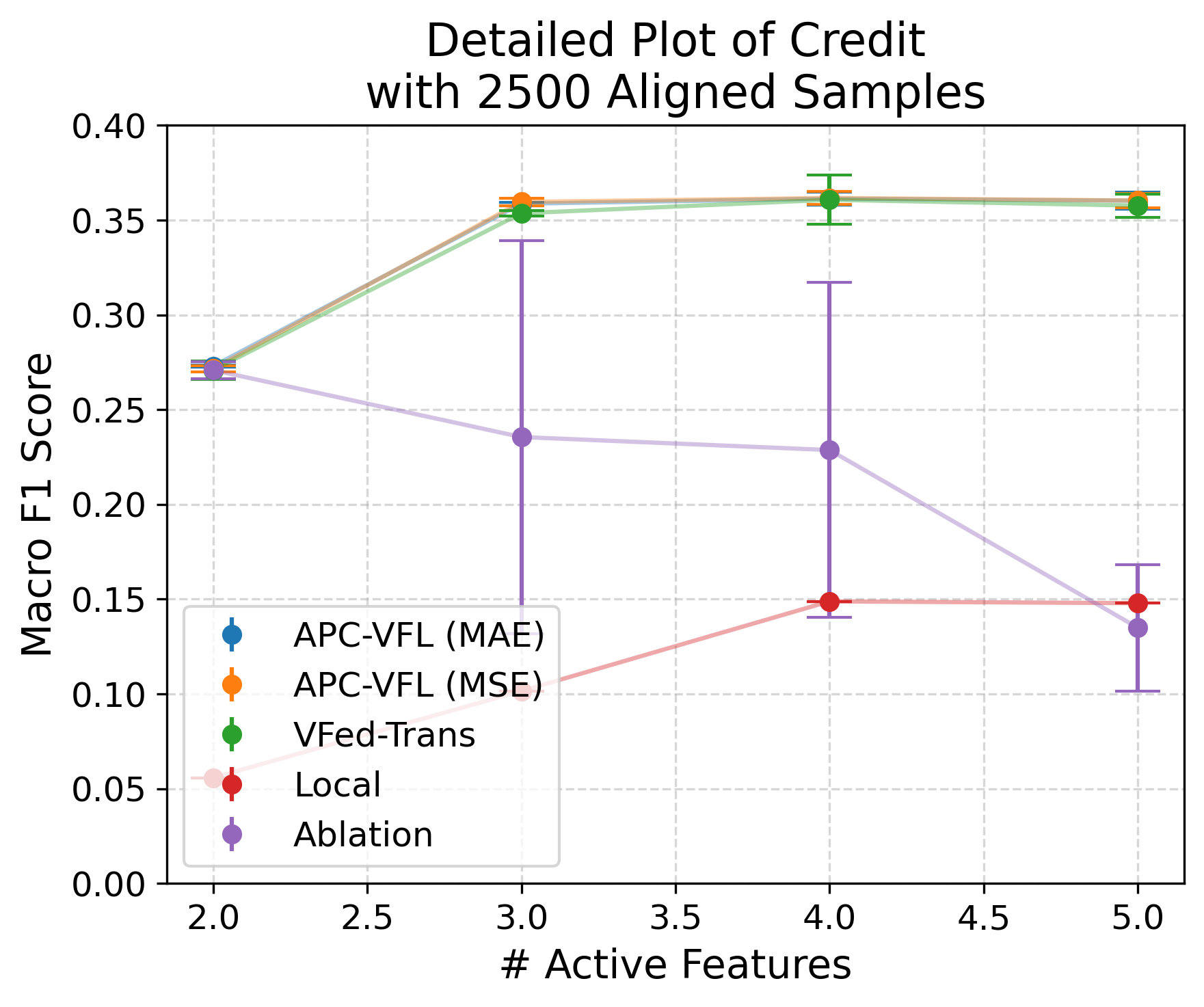}
    }
    \hspace{0.01\textwidth}
    \subfigure[Detailed plot of UCI Credit Card with 5000 aligned samples.]{%
        \includegraphics[width=0.28\textwidth]{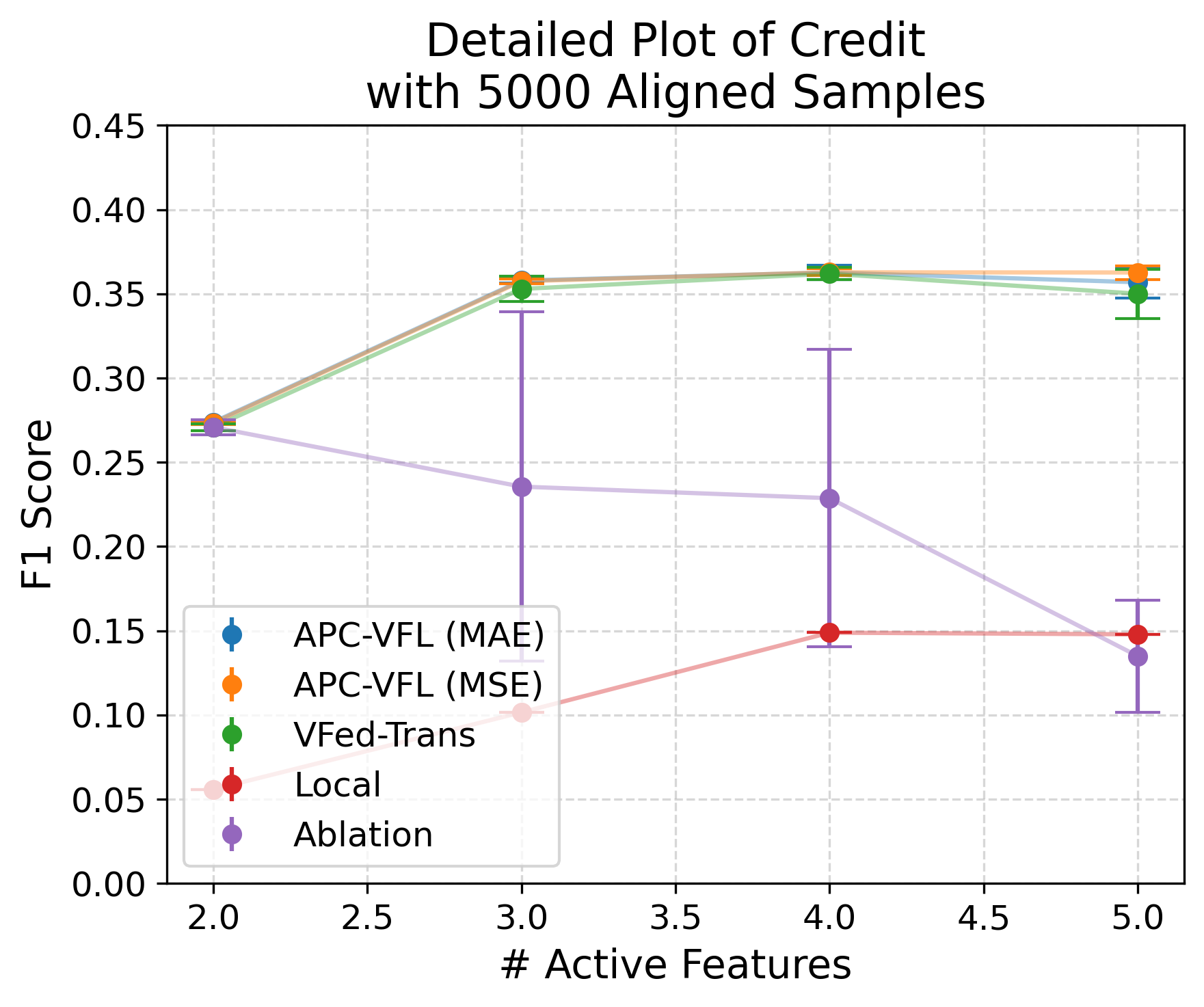}
    }
    \hspace{0.01\textwidth}
    \subfigure[Detailed plot of UCI Credit Card with 7500 aligned samples.]{%
        \includegraphics[width=0.28\textwidth]{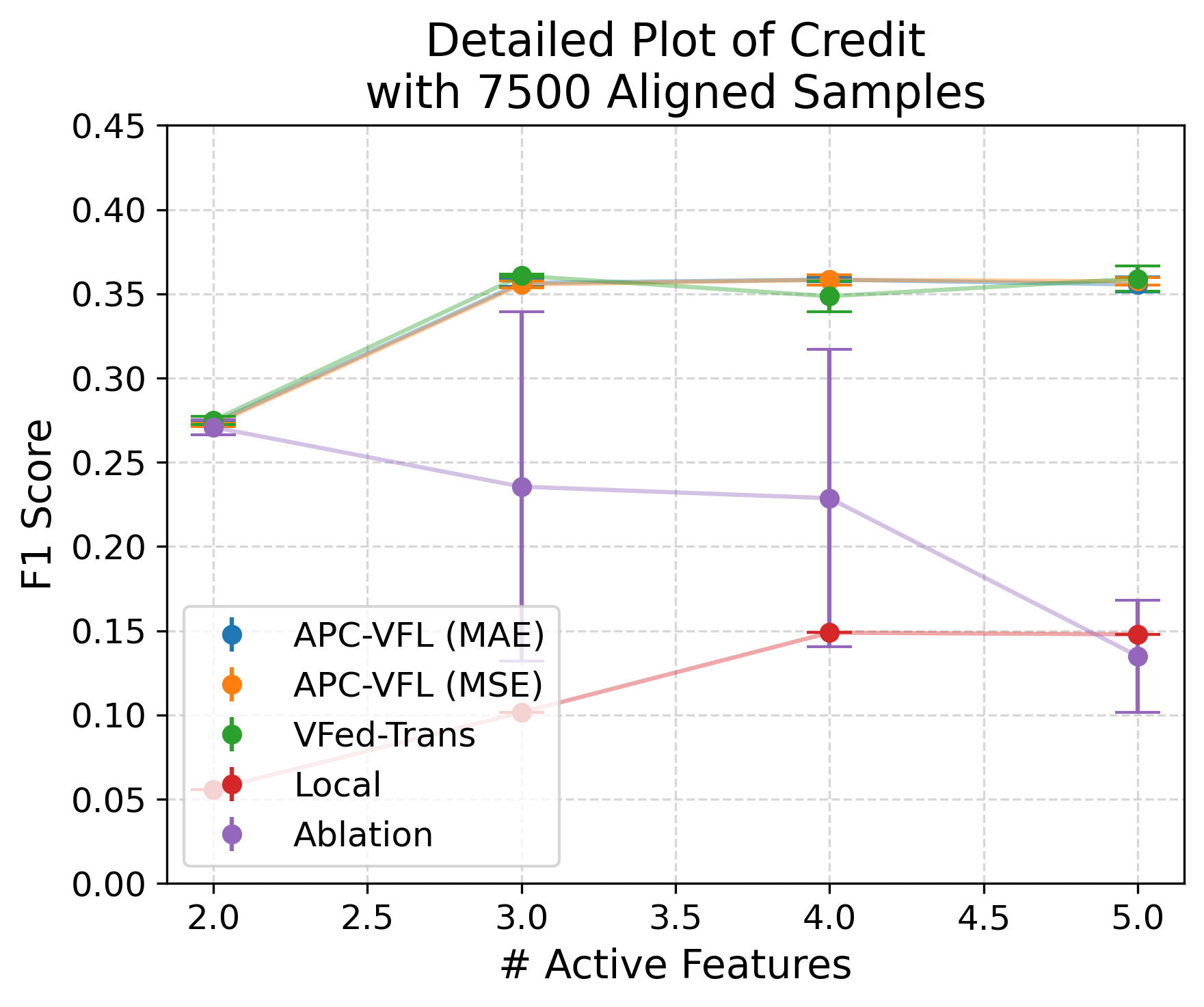}
    }
    \hspace{0.01\textwidth}
    \subfigure[Detailed plot of UCI Credit Card with 10000 aligned samples.]{%
        \includegraphics[width=0.28\textwidth]{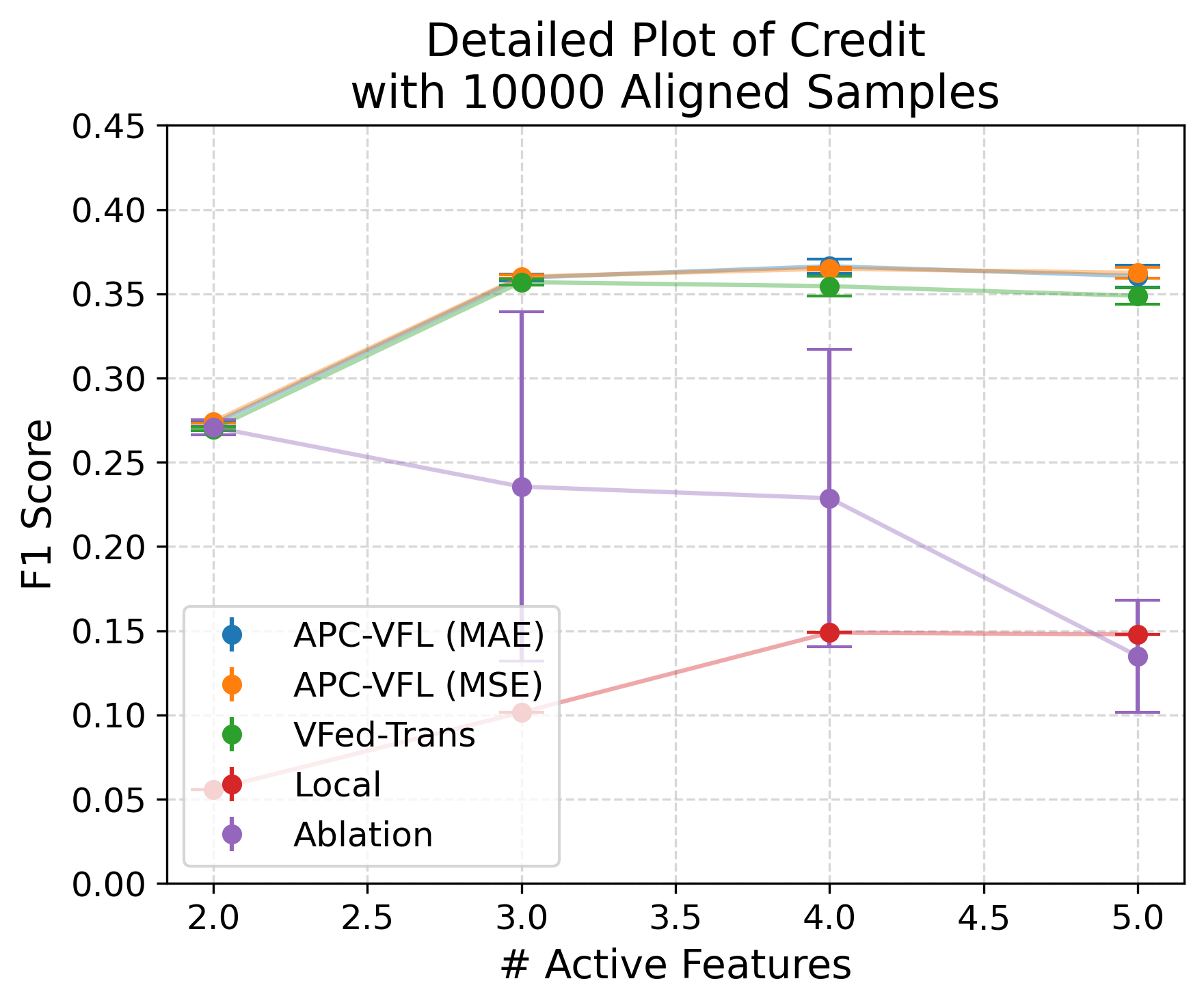}
    }
    \caption{Detailed plots for UCI Credit Card with varying sample alignment.}
    \label{fig:fine grained credit}
\end{figure}

\section{Extra results}\label{app: extra results}
\subsection{Macro and weigted averaged F1 scores for MIMIC-III}
\begin{figure}[h]
  \centering
  \includegraphics[width=0.95\textwidth]{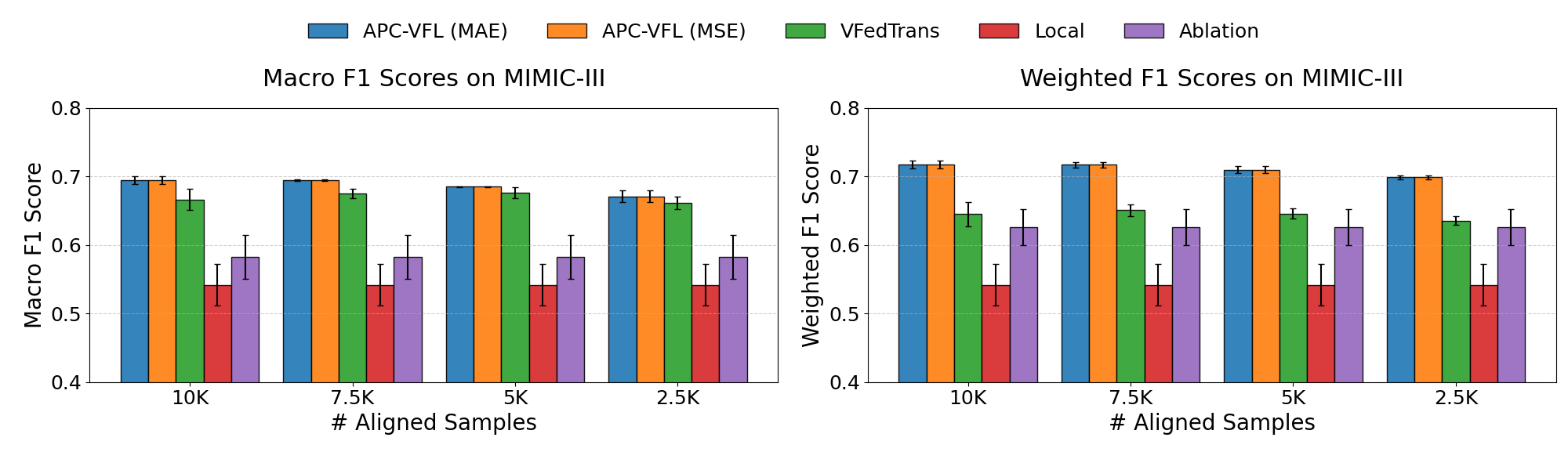}
  \caption{Mean results on MIMIC-III with \textit{macro} and \textit{weighted} averaged F1 score with varying aligned samples.}
  \label{fig:micro macro mimic}
\end{figure}

The mean results with \textit{macro} and \textit{weighted} averaged F1 score on MIMIC-III with limited sample overlap are shown in Figure \ref{fig:micro macro mimic}.

\begin{figure}[h]
    \centering
    \subfigure[Fine-grained plot of MIMIC-III with 2500 aligned samples and \textit{macro} averaging.]{%
        \includegraphics[width=0.28\textwidth]{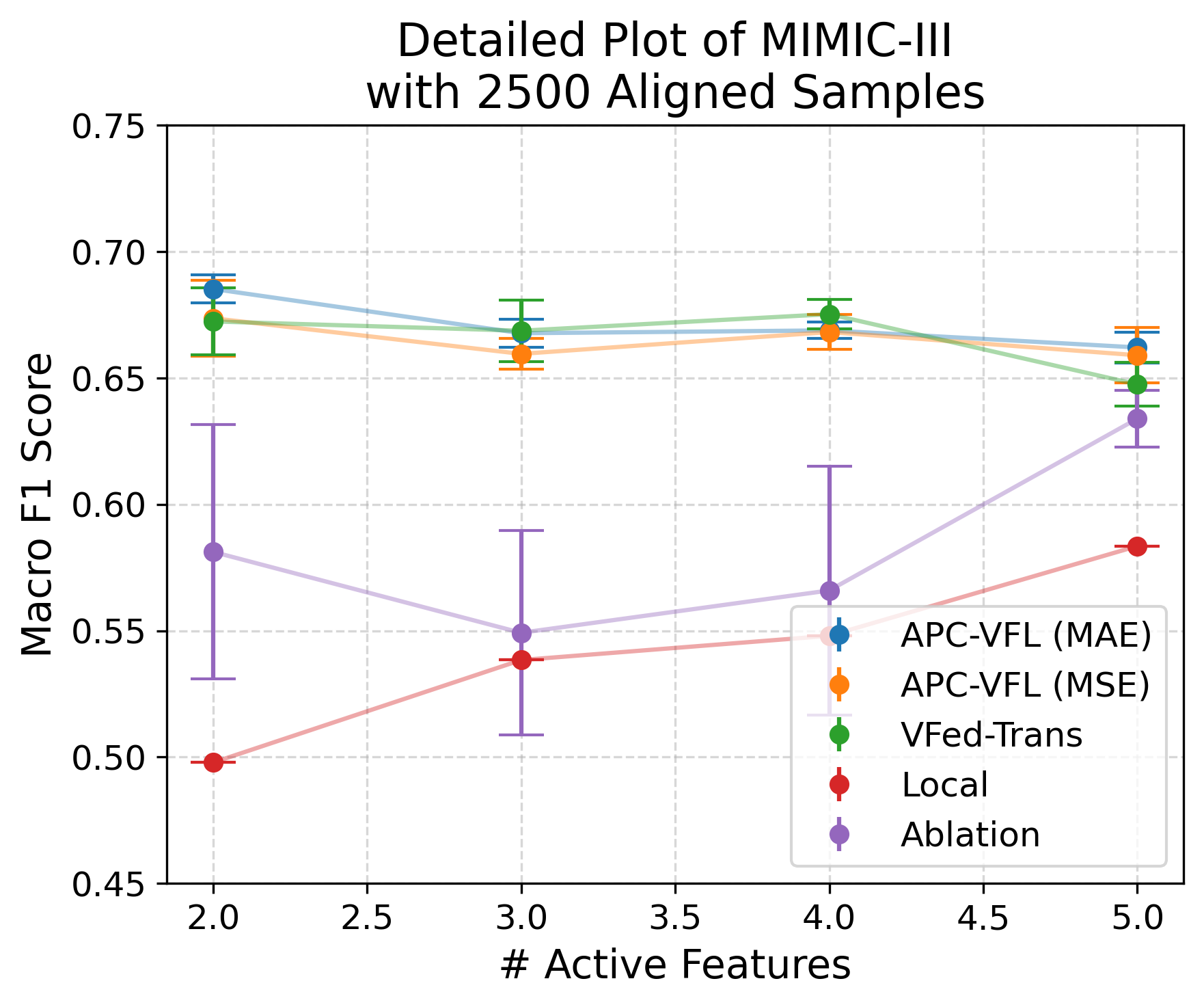}
    }
    \hspace{0.01\textwidth}
    \subfigure[Fine-grained plot of MIMIC-III with 5000 aligned samples and \textit{macro} averaging.]{%
        \includegraphics[width=0.28\textwidth]{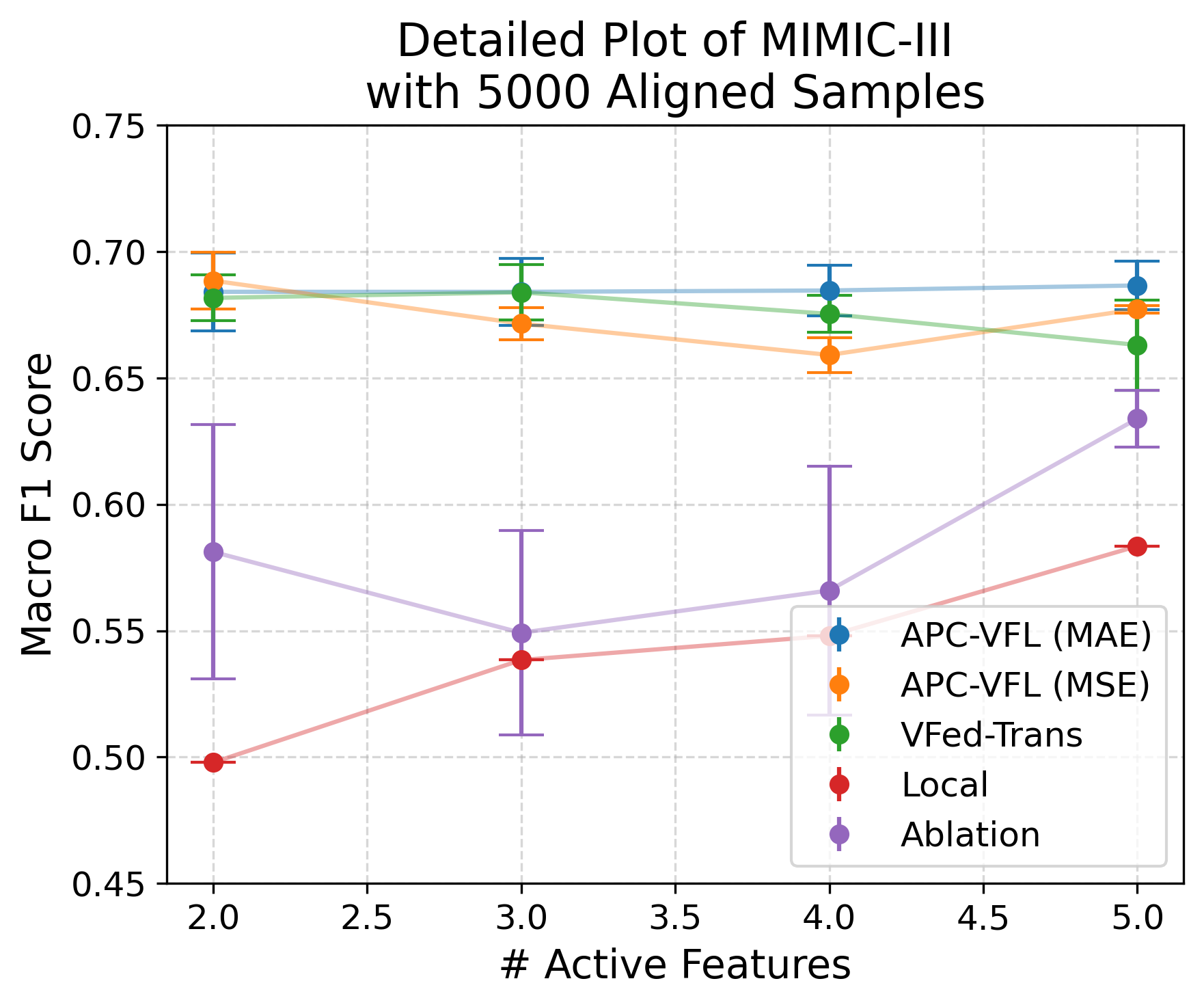}
    }
    \hspace{0.01\textwidth}
    \subfigure[Fine-grained plot of MIMIC-III with 7500 aligned samples and \textit{macro} averaging.]{%
        \includegraphics[width=0.28\textwidth]{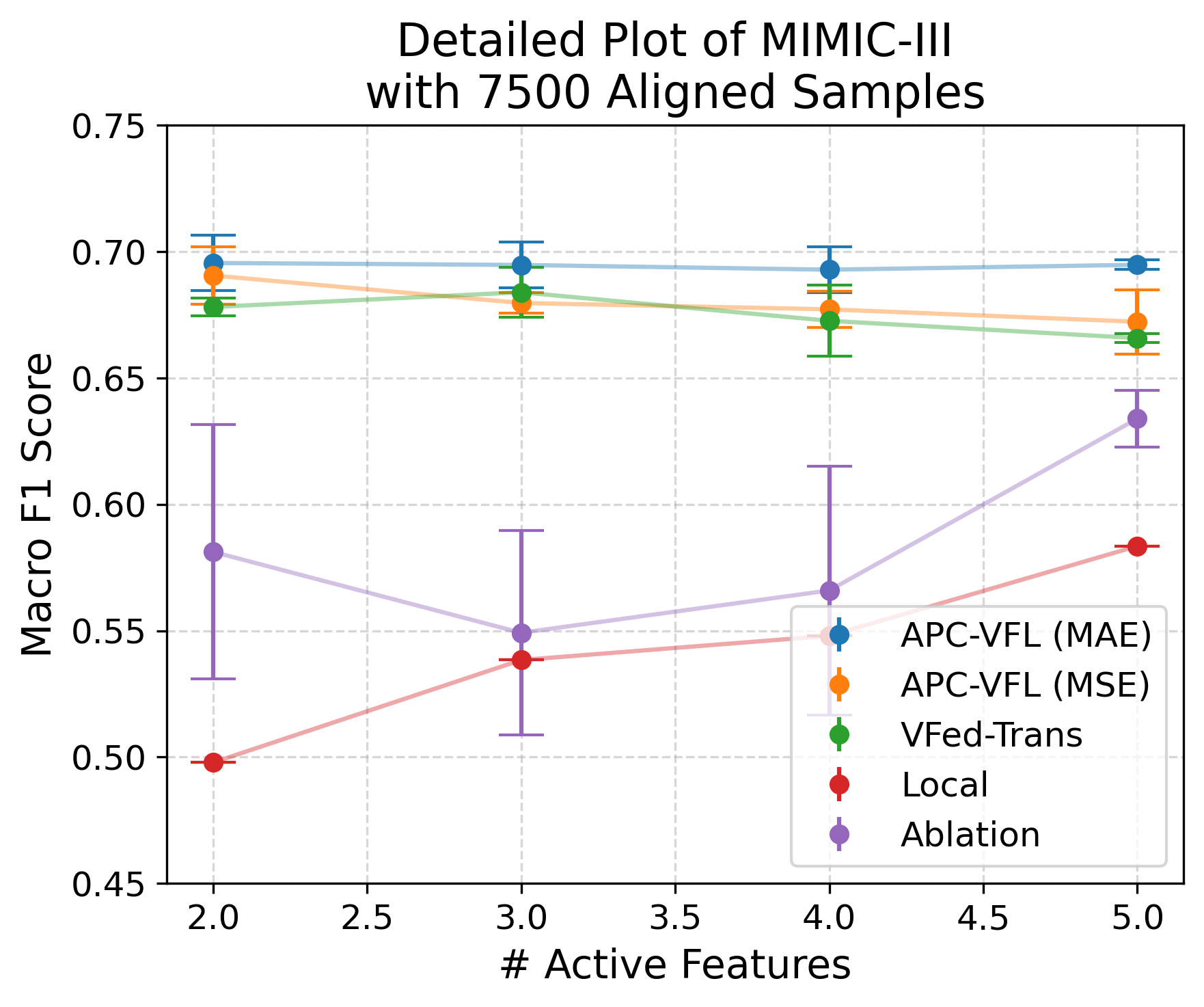}
    }
    \hspace{0.01\textwidth}
    \subfigure[Fine-grained plot of MIMIC-III with 10000 aligned samples and \textit{macro} averaging.]{%
        \includegraphics[width=0.28\textwidth]{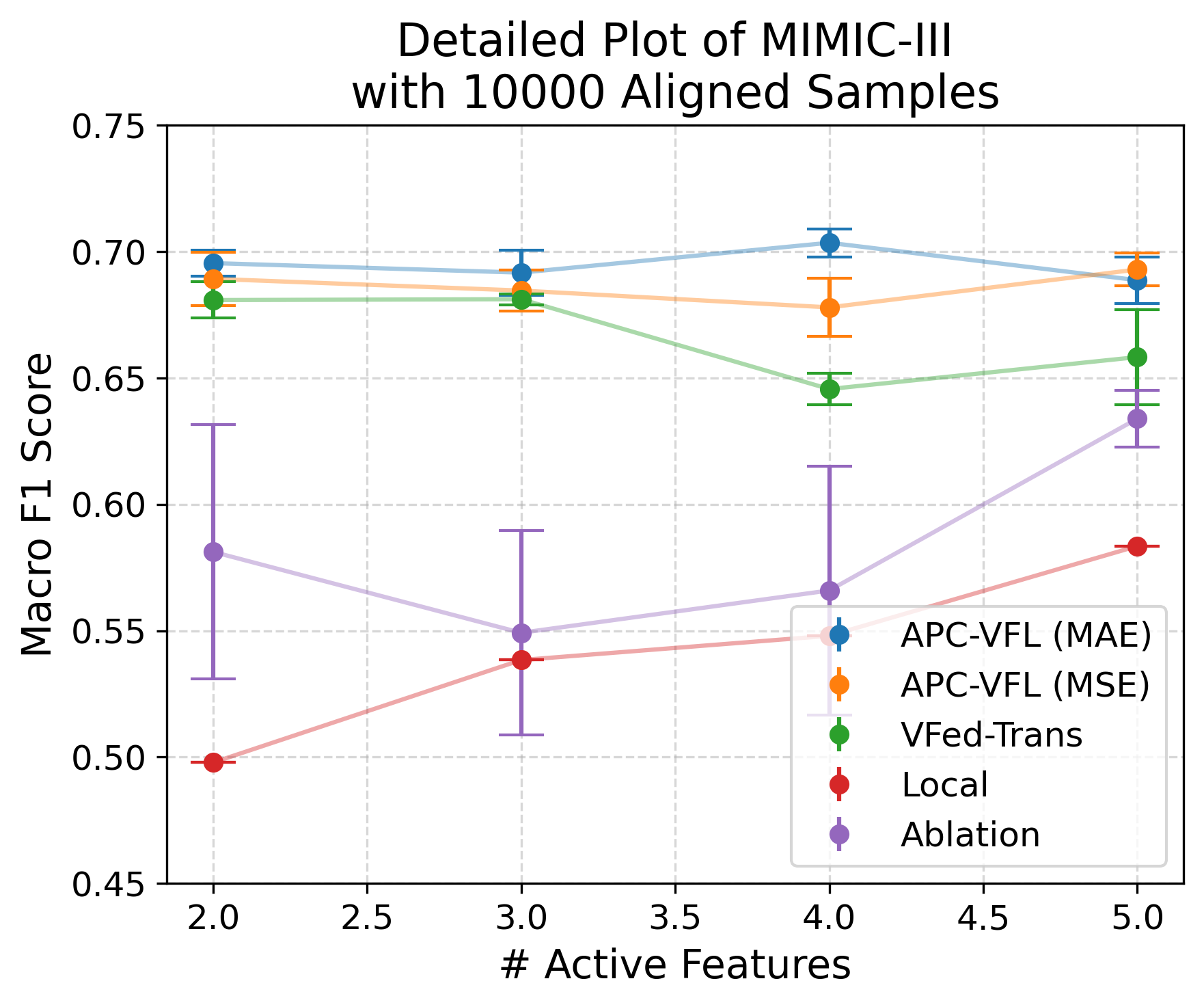}
    }
    \caption{Fine-grained plots for MIMIC-III with varying sample alignment and \textit{macro} averaged F1.}
    \label{fig:fine grained MIMIC macro}
\end{figure}

\begin{figure}[h]
    \centering
    \subfigure[Fine-grained plot of MIMIC-III with 2500 aligned samples and \textit{weighted} averaging.]{%
        \includegraphics[width=0.27\textwidth]{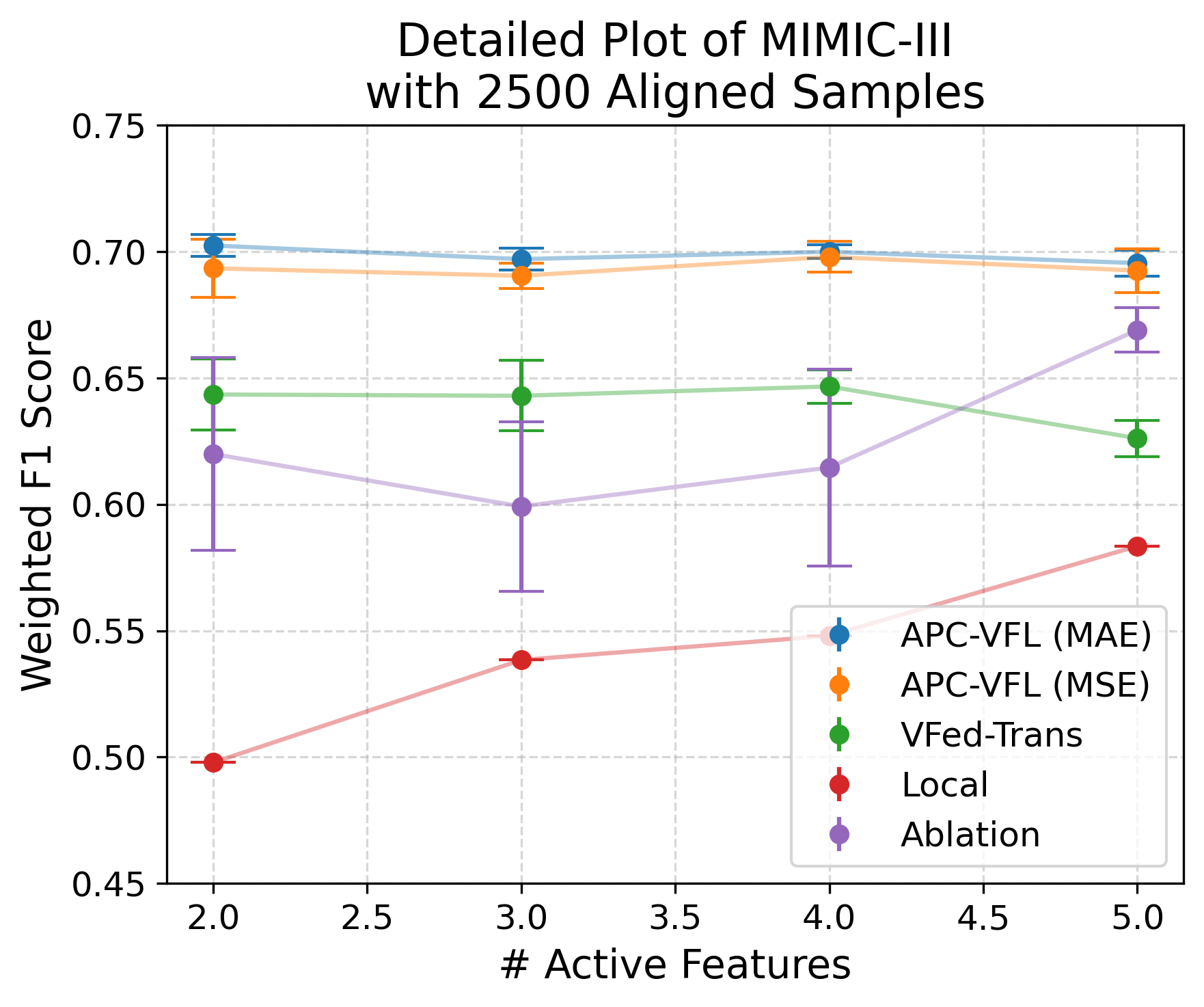}
    }
    \hspace{0.05\textwidth}
    \subfigure[Fine-grained plot of MIMIC-III with 5000 aligned samples and \textit{weighted} averaging.]{%
        \includegraphics[width=0.27\textwidth]{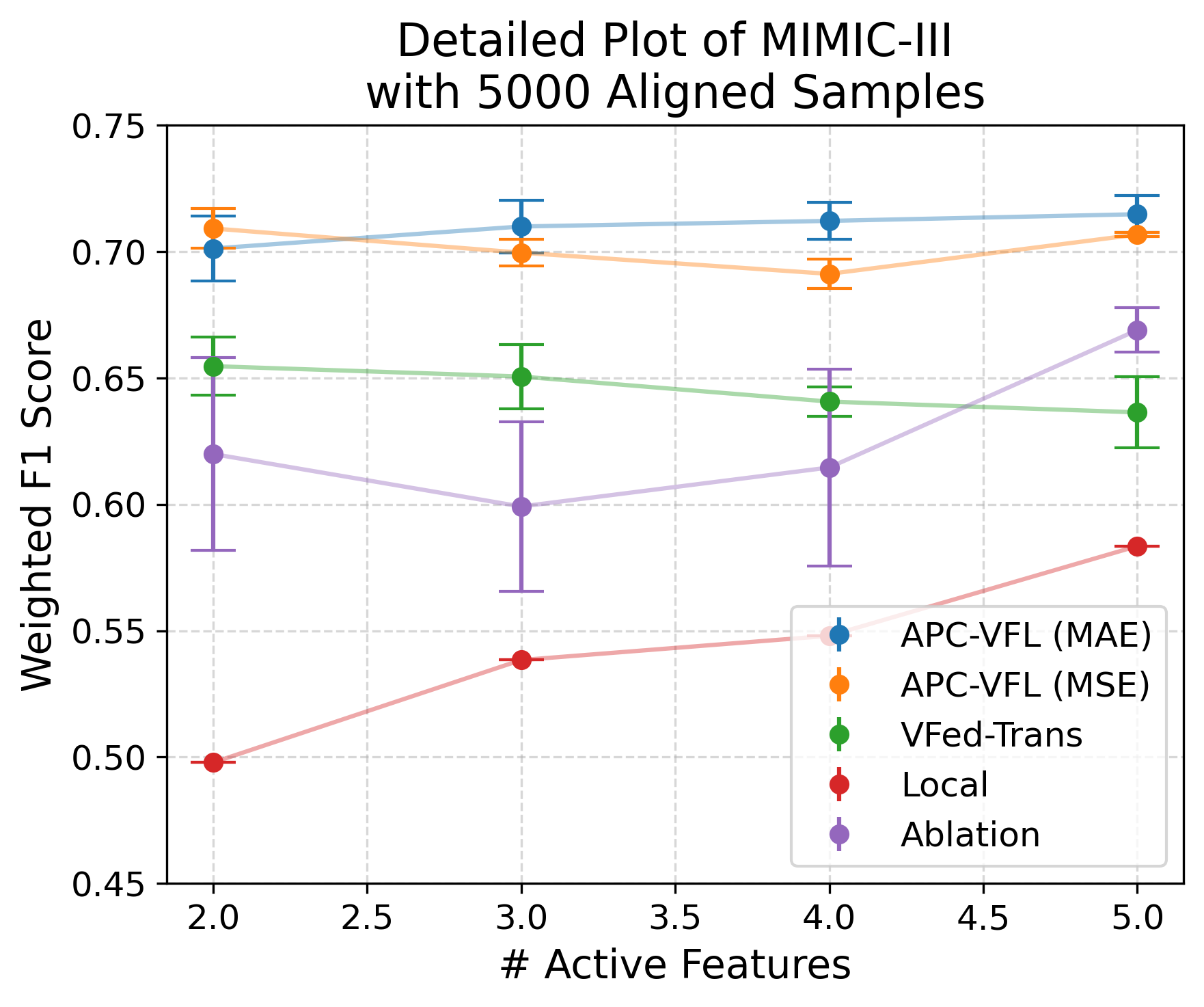}
    }
    \hspace{0.05\textwidth}
    \subfigure[Fine-grained plot of MIMIC-III with 7500 aligned samples and \textit{weighted} averaging.]{%
        \includegraphics[width=0.27\textwidth]{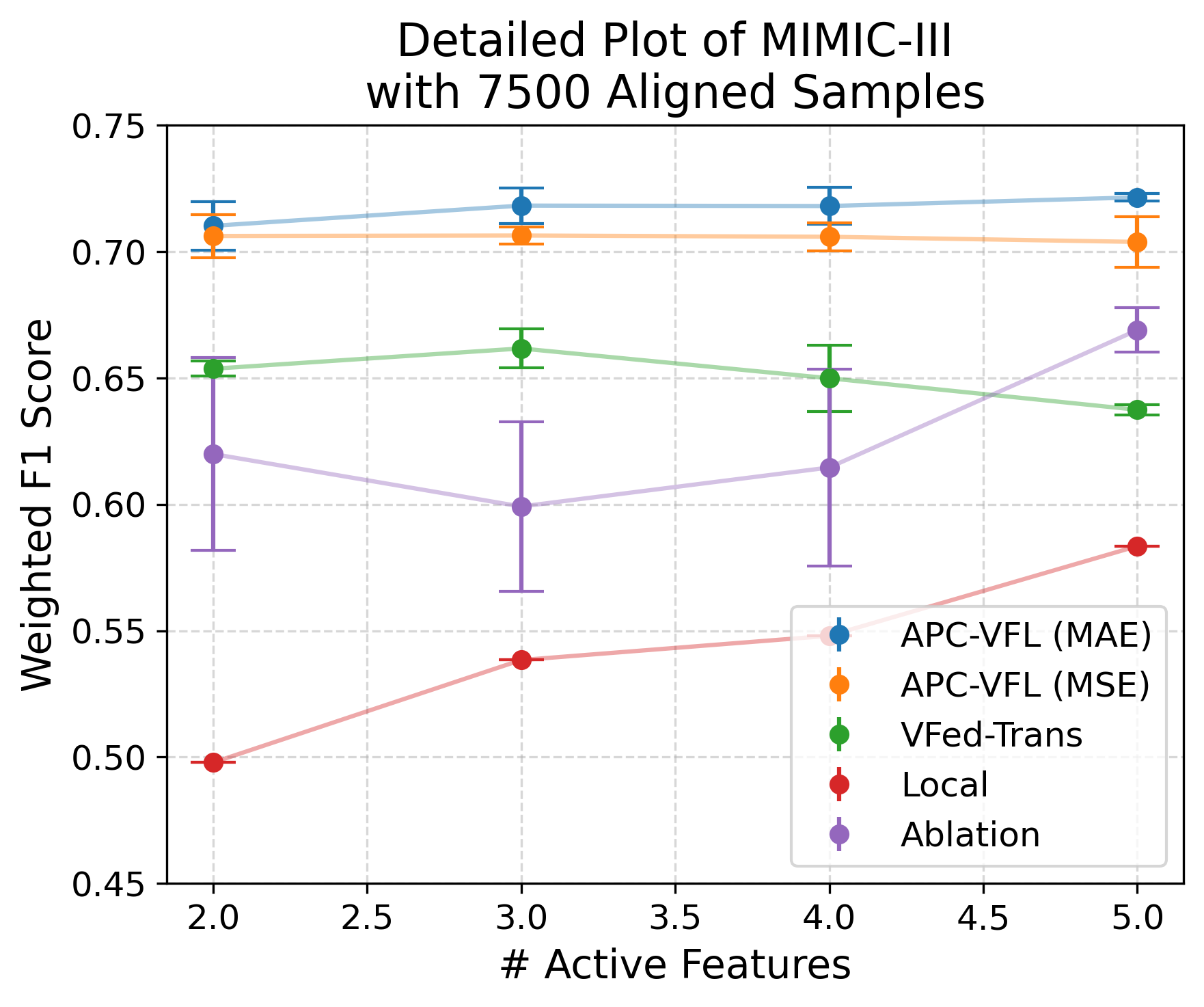}
    }
    \hspace{0.05\textwidth}
    \subfigure[Fine-grained plot of MIMIC-III with 10000 aligned samples and \textit{weighted} averaging.]{%
        \includegraphics[width=0.27\textwidth]{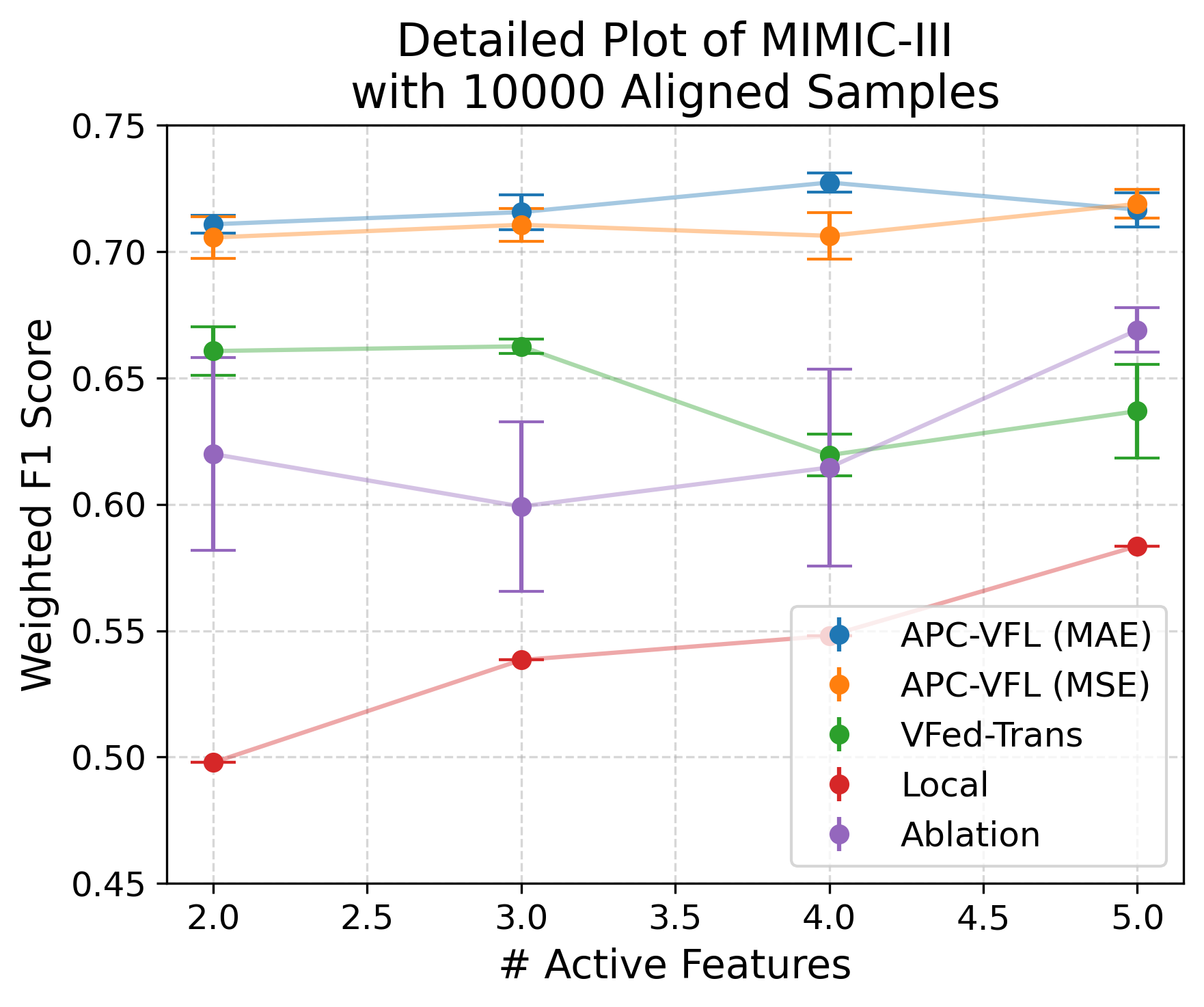}
    }
    \caption{Fine-grained plots for MIMIC-III with varying sample alignment and \textit{weighted} averaged F1.}
    \label{fig:fine grained MIMIC micro}
\end{figure}

\newpage
As previously done, we also show the detailed decomposition of the plots from Figure \ref{fig:micro macro mimic}:

Figure \ref{fig:fine grained MIMIC macro} shows the fine grained plots of the \textit{macro} averaged F1 scores.
\newpage
Detailed results of the \textit{weighted} averaged F1 scores are show in Figure \ref{fig:fine grained MIMIC micro}.

The \textit{macro} and \textit{weighted} averaged values of the comparison with Vertical SplitNN are presented on Table \ref{tab: macro weighted mimic splitNN apc-vfl}:

\begin{table}[h]
\centering
\caption{\textit{Macro} and \textit{weighted} averaged F1 scores of SplitNN and APC-VFL on MIMIC-III.}
\label{tab: macro weighted mimic splitNN apc-vfl}
\begin{tabular}{llcc}
\toprule
\textbf{\# Aligned}   & \textbf{Method} & \textbf{Macro F1} & \textbf{Weighted F1} \\ \hline
\multirow{2}{*}{10K}  & SplitNN         & $81.46 \scalebox{0.9}{\(\pm 0.41\)}$     & $82.39 \scalebox{0.9}{\(\pm 0.24\)}$          \\
                      & APC-VFL         & $70.02 \scalebox{0.9}{\(\pm 0.60\)}$      & $72.80 \scalebox{0.9}{\(\pm0.45\)}$          \\ \hline
\multirow{2}{*}{7.5K} & SplitNN         & $73.41 \scalebox{0.9}{\(\pm4.36\)}$     & $76.52 \scalebox{0.9}{\(\pm3.73\)}$          \\
                      & APC-VFL         & $68.95 \scalebox{0.9}{\(\pm0.26\)} $     & $73.50 \scalebox{0.9}{\(\pm0.31\)}$          \\ \hline
\multirow{2}{*}{5K}   & SplitNN         & $68.85 \scalebox{0.9}{\(\pm0.94\)} $     & $72.01\scalebox{0.9}{\(\pm0.61\)}$          \\
                      & APC-VFL         & $69.21 \scalebox{0.9}{\(\pm1.23\)} $     & $72.90 \scalebox{0.9}{\(\pm1.01\)}$          \\ \hline
\multirow{2}{*}{2.5K} & SplitNN         & $62.58 \scalebox{0.9}{\(\pm0.79\)}$       & $67.0 \scalebox{0.9}{\(\pm0.57\)}$          \\
                      & APC-VFL         & 67.14 \scalebox{0.9}{\(\pm0.48\)}      & $70.95 \scalebox{0.9}{\(\pm0.47\)}$          \\ 
\bottomrule
\end{tabular}
\end{table}

The results obtained with different averaging methods align with prior analysis.

\subsection{Macro and weighted averaged results on MIMIC-III with fewer aligned samples}

\begin{figure}[h]
  \centering
  \includegraphics[width=0.95\textwidth]{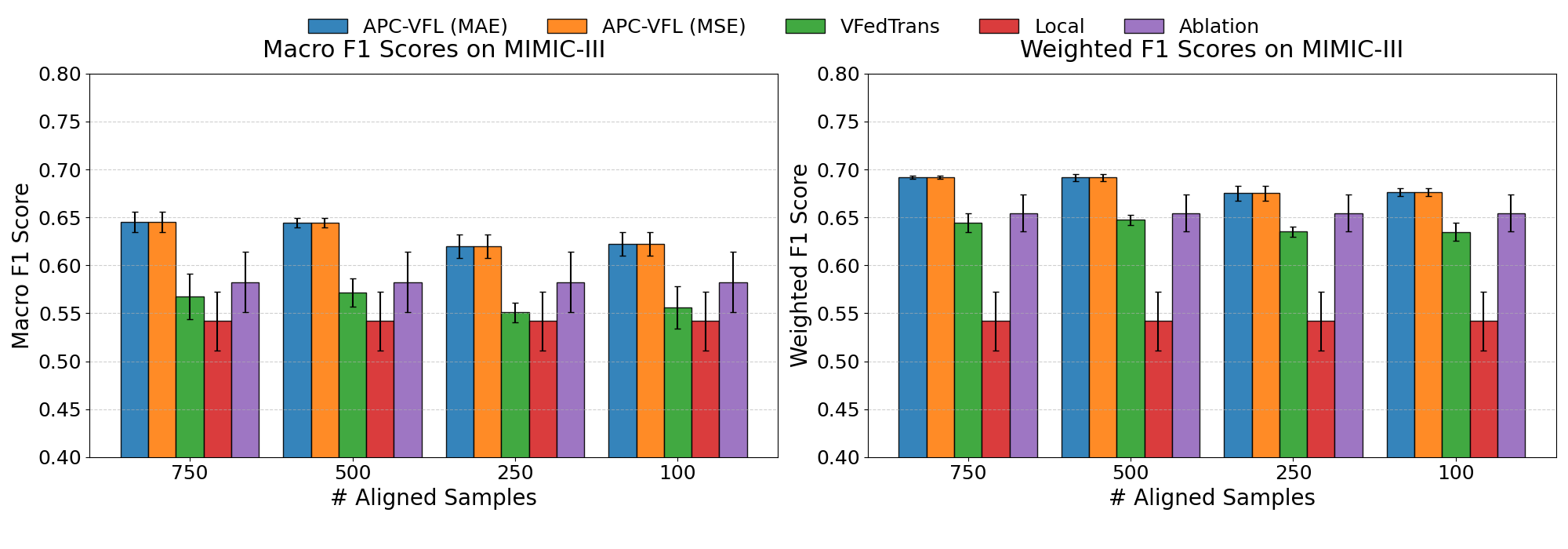}
  \caption{Mean results on MIMIC-III with minimum sample overlap.}
  \label{fig:mimic minimum overlap}
\end{figure}

Figure \ref{fig:mimic micro F1 minimum overlap} shows the F1 scores with \textit{micro} averaging on MIMIC-III with \{750, 500, 250, 100\} aligned samples. The corresponding results for \textit{macro} and \textit{weighted} averaging are shown in Figure \ref{fig:mimic minimum overlap}, further supporting the earlier findings.

\subsection{Results on UCI Credit Card with fewer aligned samples}
\begin{figure}[h]
  \centering
  \includegraphics[width=0.55\textwidth]{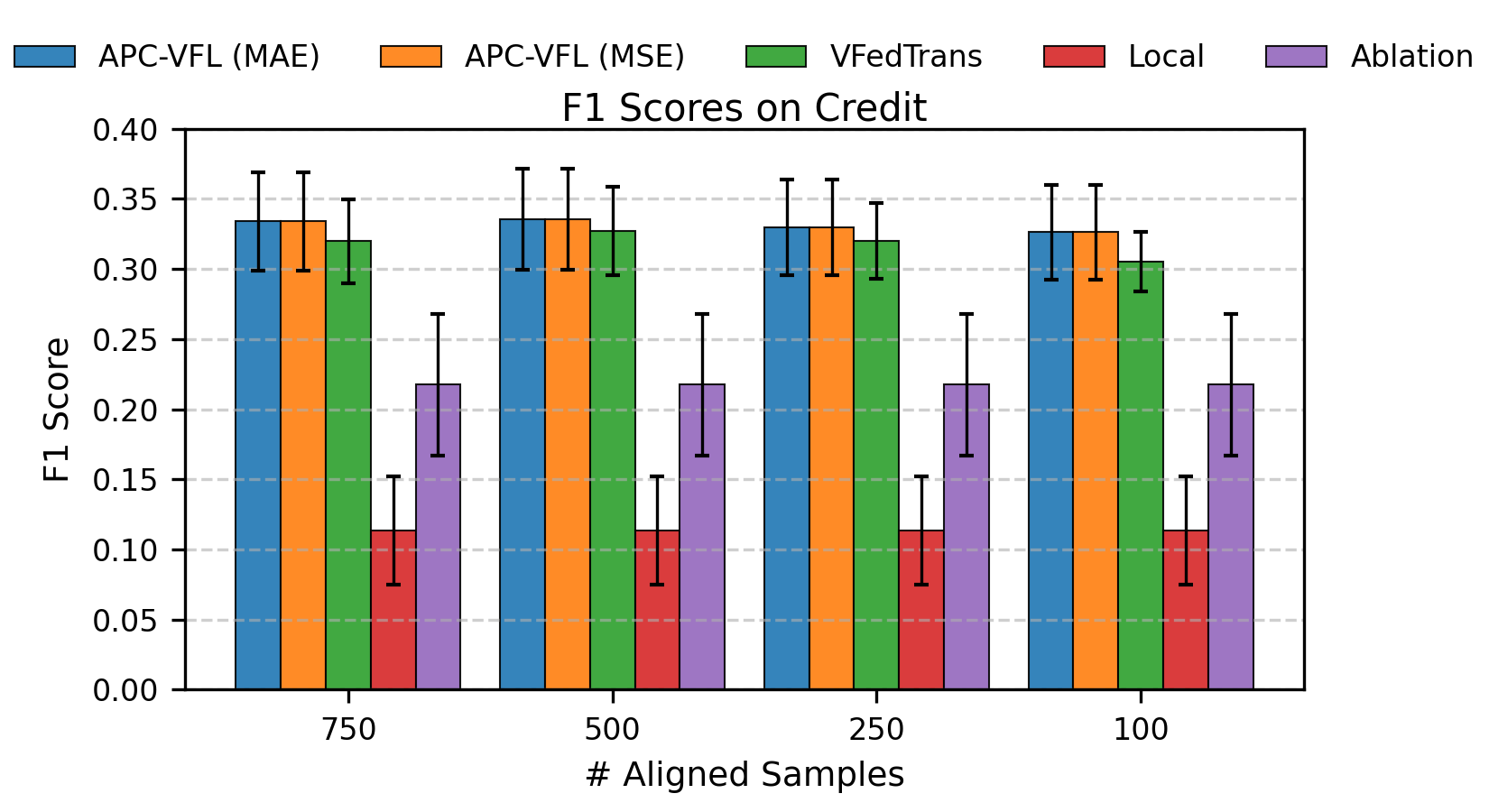}
  \caption{Mean results on UCI Credit Card with minimum sample overlap.}
  \label{fig:UCI credit minimum overlap}
\end{figure}

Even if the performance decay on UCI Credit Card dataset is not as pronounced as on MIMIC-III, we have also performed the experiments further reducing the aligned sample number on this dataset. The results with $750, 500, 250\; \text{and}\; 100$ aligned samples can found on Figure \ref{fig:UCI credit minimum overlap}.

It can be observed that the performance decay is minimal on this scenario as well with APC-VFL. Our approach has a degradation of $\approx1.3$ points in F1 score on average between the scenario with 10000 aligned samples, shown in Figure \ref{fig: results federated present}, and the scenario with 100 aligned samples that can be seen on Figure \ref{fig:UCI credit minimum overlap}. On the other hand, the degradation of VFedTrans doubles the one of our method, as it losses $\approx2.7$ points of F1 on average.

\section{Communication footprint computation}
On this Appendix we provide the expressions that have been used to compute the communication footprint of the different methods.

\subsection{APC-VFL}
Our proposal requires a single information exchange. On this exchange a data matrix $Z_A\in\mathbb{R}^{|\mathcal{D}_A|\times z_p}$ is shared, where $z_p$ is the dimensions of passive participant's embeddings.

Since each one of the elements of $Z_A$ weights 4 bytes, the total footprint can be computed as follows:
\begin{equation}
\label{equ: footprint apc-vfl}
    \text{footprint}=|\mathcal{D}_A|\cdot z_p \cdot 4\; \text{(bytes)}
\end{equation}

In our experiments, we have set $z_p=256$, as can be seen on Table \ref{tab:architecture_details}. This is the reason why the communication overhead of our proposal increases linearly with the aligned samples. Anyway, it must be noted that even if $|\mathcal{D}_A|$ is defined by the VFL scenario, by adjusting $z_p$ the footprint can also be adjusted.

\subsection{SplitNN}
When computing the communication cost of SplitNN, the forward pass as well as the backpropagation must be taken into account.

\paragraph{Forward pass}
The communication footprint of performing the forward pass of the whole dataset can be computed with Equation \ref{equ: footprint apc-vfl}. Given that this process is done as many times as training epochs, the total communication cost of the forward passes is:

\begin{equation}
\label{equ: forward splitNN}
    \text{forward} = \text{epochs}\cdot |\mathcal{D}_A|\cdot z_p \cdot 4\; \text{(bytes)}
\end{equation}

\paragraph{Backpropagation}
To backpropagate the error through the local feature extractor of the passive participant, the information being sent from the active has as many elements as parameters of the final layer of the local passive encoder. Being the number of parameters $p$, the communication footprint is the following one:

\begin{equation}
\label{equ: backprop splitNN}
    \text{backprop} = \text{epochs}\cdot\left\lceil\frac{|\mathcal{D}_A|}{\text{Batch size}}\right\rceil\cdot p\cdot4\; \text{(bytes)}
\end{equation}

Taking into account that the final layer of the local passive encoder has an input dimension of $128$ and an output of $256$ neurons, the total number of parameters is: $p=128\cdot256 + 256$.

Notice how ``$\text{epochs}\cdot\left\lceil\frac{|\mathcal{D}_A|}{\text{Batch size}}\right\rceil$'' is the total number of times backpropagation is done. Therefore, the total communication rounds shown on Table \ref{tab:merged_results}, doubles this number.

\paragraph{Total communication footprint}
The total communication footprint is nothing but the sum of the footprint of the \textit{forward pass} (Equation \ref{equ: forward splitNN}) and the \textit{backpropagation} (Equation \ref{equ: backprop splitNN}):

\begin{equation}
    \text{footprint} = \text{forward} + \text{backprop}
\end{equation}

\paragraph{VFedTrans}
For VFedTrans, the chosen representation learning process has been FedSVD, as it performs the best. This process requires a total of five information exchanges where the following matrices are shared:

\begin{equation*}
    A\in\mathbb{R}^{|\mathcal{D}_A|\times|\mathcal{D}_A|};\; B_t\in\mathbb{R}^{x_t\times x_{total}};\;B_d\in\mathbb{R}^{x_d\times x_{total}};\;\tilde{S}_t\in\mathbb{R}^{|\mathcal{D}_A|\times x_t};\; \tilde{S}_d\in\mathbb{R}^{|\mathcal{D}_A|\times x_d};\; \tilde{U}\in\mathbb{R}^{|\mathcal{D}_A|\times x_{total}}
\end{equation*}

where,
\begin{description}
    \item $x_t$ is the number of features of active participant's dataset.
    \item $x_d$ is the number of features of the passive participant's dataset.
    \item $x_{total} = x_t + x_d$
\end{description}

Notice how $A$ is a square matrix with as many rows and columns as the aligned samples. This matrix is the reason for the exponential growth of the communication footprint shown in Figure \ref{fig:comm mimic}.

From the information exchange point of view, the process is the following one:
\begin{itemize}
    \item A trusted key generator sends $A$ and $B_t$ to the active participant.
    \item A trusted key generator sends $A$ and $B_d$ to the passive participant.
    \item The active participant sends $\tilde{S}_t$ to a third party server.
    \item The passive participant sends $\tilde{S}_d$ to a third party server.
    \item The third party server sends $\tilde{U}$ to the active participant.
\end{itemize}

Taking these five information exchanges into account, the communication cost can be computed by computing the weight of each one of the elements that are being sent.

\begin{equation}
    \text{footprint} = (2\cdot|\mathcal{D}_A|^2+x_t\cdot x_{total} + x_d\cdot x_{total}+|\mathcal{D}_A|\cdot x_t + |\mathcal{D}_A|\cdot x_d+|\mathcal{D}_A|\cdot x_{total})\cdot 4\;\text{(bytes)} 
\end{equation}

\section{Method to train encoders while evaluating the quality of representations}
The proposed approach of training multiple encoders and a classifier introduces a vast and complex search space for selecting architectures and training procedures. Exploring this search space through full federated learning processes is computationally prohibitive and time consuming, creating a critical need for a practical evaluation mechanism to assess each encoder individually. To address this challenge, we propose a method for training encoders that evaluates the quality of their representations during the training process. This method provides a foundation for making informed decisions about encoder architectures and training techniques by offering insights into how different choices affect representation quality. Importantly, this approach is not intended for use during the actual federated learning process but rather as a tool for optimizing design decisions beforehand.

Considering that the ultimate goal of the federated process is to train a machine learning model based on data representations rather than the original variables, it is essential to evaluate the different encoder architectures and training methods accordingly. We propose training the encoders following Algorithm \ref{algo: encoder training} to assess the effectiveness of various architectures and training approaches.

\begin{algorithm}
\caption{Algorithm to train an encoder evaluating the quality of the representations}
\label{algo: encoder training}
\begin{algorithmic}[1]
\STATE \textbf{Input:} Classifier $C$
\STATE \textbf{Input:} Classification quality evaluation metric $m$ (\textit{F1, accuracy, etc.}), to get sets of metrics $\tilde{\mathcal{M}}$
\STATE \textbf{Input:} Training data in feature space $X$, with corresponding labels $\mathbf{y}$
\STATE \textbf{Input:} Encoder $g$
\STATE \textbf{Input:} Loss function for training, \textit{loss}
\STATE \textbf{Input:} Amount of training epochs, \textit{epochs}
\STATE \textbf{Output:} Training \textit{loss} and representation quality evolution based on $m$
\FOR{\textit{epoch} in \textit{epochs}}
    \STATE Train the encoder with $X$
    \STATE Compute and store epoch \textit{loss}
    \STATE Obtain $Z$ from $X$ using the encoder
    \STATE Perform a $k$-fold cross validation on $Z$ using the labels $\mathbf{y}$ and the defined classifier $\mathcal{C}$
    \STATE Evaluate the classification results of each fold with $m$, obtaining the set $\tilde{\mathcal{M}}$
\ENDFOR
\STATE \textbf{Return:} Evolution of the training loss and sets of metrics, $\tilde{\mathcal{M}}$, per \textit{epoch}
\end{algorithmic}
\end{algorithm}

Following the algorithm, a set of metrics based on the learned data representations, denoted as $\tilde{\mathcal{M}}$, is obtained. Note that $|\tilde{\mathcal{M}}|=k$, as a value is obtained per fold. However, these metrics alone do not provide sufficient information about the quality of the representations. To address this, we perform another $k$-fold cross-validation using a new instance of the same classifier, $\mathcal{C}$, on the original data, $X$, yielding a new set of metrics, denoted as $\mathcal{M}$. Based on the metrics from both the original data and its representations, we make the following assumption:

\begin{equation}
    \frac{\sum_{i\in\mathcal{M}}i}{\lvert\mathcal{M}\rvert} \approx \frac{\sum_{i\in\mathcal{\tilde{\mathcal{M}}}}i}{\lvert\tilde{\mathcal{M}}\rvert}  \implies X \approx Z
\end{equation}

This implies that in terms of information, the original variables and the representations are equivalent if the sets of metrics are similar. To be more precise, we define the sets as similar if the difference between their means is smaller than a predetermined threshold, $r$:

\begin{equation}
    \frac{\sum_{i\in\mathcal{M}}i}{\lvert\mathcal{M}\rvert} - \frac{\sum_{i\in\mathcal{\tilde{\mathcal{M}}}}i}{\lvert\tilde{\mathcal{M}}\rvert} \le r \implies X \approx Z
\end{equation}

The choice of metric and the threshold $r$ are application-dependent, but this methodology provides a structured way to assess the quality of different encoder architectures and training methods. Consequently, this approach has been employed to guide decisions regarding these aspects of the proposed framework.


\end{document}